\documentclass[11pt, a4paper, logo]{googledeepmind}

\usepackage[authoryear, sort&compress, round]{natbib}
\usepackage{algorithm}
\usepackage{algpseudocode}

\newcommand{\beginsupplement}{%
        \setcounter{table}{0}
        \renewcommand{\thetable}{S\arabic{table}}%
        \setcounter{figure}{0}
        \renewcommand{\thefigure}{S\arabic{figure}}%
     }

\title{Graph schemas as abstractions for transfer learning, inference, and planning}
\correspondingauthor{dileepgeorge@deepmind.com}

\author[1]{J. Swaroop Guntupalli}
\author[1]{Rajkumar Vasudeva Raju}
\author[1]{Shrinu Kushagra}
\author[1]{Carter Wendelken}
\author[1]{Danny Sawyer}
\author[2]{Ishan Deshpande}
\author[1]{Guangyao Zhou}
\author[1]{Miguel L\'{a}zaro-Gredilla}
\author[1]{Dileep George}

\affil[1]{Google DeepMind}
\affil[2]{Google Research}

\begin{abstract}
Transferring latent structure from one environment or problem to another is a mechanism by which humans and animals generalize with very little data. Inspired by cognitive and neurobiological insights, we propose graph schemas as a mechanism of abstraction for transfer learning. Graph schemas start with latent graph learning where perceptually aliased observations are disambiguated in the latent space using contextual information. Latent graph learning is also emerging as a new computational model of the hippocampus to explain map learning and transitive inference. Our insight is that a latent graph can be treated as a flexible template — a schema — that models concepts and behaviors, with slots that bind groups of latent nodes to the specific observations or groundings. By treating learned latent graphs (schemas) as prior knowledge, new environments can be quickly learned as compositions of schemas and their newly learned bindings. We evaluate graph schemas on two previously published challenging tasks: the memory \& planning game and one-shot StreetLearn, which are designed to test rapid task solving in novel environments. Graph schemas can be learned in far fewer episodes than previous baselines, and can model and plan in a few steps in novel variations of these tasks. We also demonstrate learning, matching, and reusing graph schemas in more challenging 2D and 3D environments with extensive perceptual aliasing and size variations, and show how different schemas can be composed to model larger and more complex environments. To summarize, our main contribution is a unified system, inspired and grounded in cognitive science, that facilitates rapid transfer learning of new environments using schemas via map-induction and composition that handles perceptual aliasing. 
\end{abstract}

\begin{document}
\maketitle
\section{Introduction}

Discovering and using the right abstractions in new situations affords efficient transfer learning as well as quick inference and planning. Humans excel at this ability, argued to be a key factor behind intelligence and a fundamental limitation in current AI systems \citep{Shanahan2022-lp}. Common reusable structured representations of concepts or behaviors---schemas---have been proposed as a powerful way to encode abstractions \citep{Mitchell2021-pc,Tenenbaum2011-cl}.
Having a computational model with the ability to discover and reuse previously-learned schemas to behave and plan in novel situations will be essential for AI.

Experimental evidence suggests that several animals have this ability \citep{farzanfar2023cognitive}. Rats and mice tend to learn new environments faster if they can reuse past schemas \citep{Tse2007-tg,Zhou2021-ja}, and macaque hippocampus cells encode spatial schemas  \citep{Baraduc2019-ro}. Neural circuits in the hippocampus and prefrontal cortex (PFC) are implicated in schema learning, recognition, update, and maintenance and these processes are considered as the foundation of memory consolidation \citep{Preston2013-pu,Gilboa2017-bk,Samborska2022-vn}. New experiences are learned in a single trial if it fits an existing schema. An updated complementary learning systems theory based on this evidence was proposed in \cite{Kumaran2016-gz}, but there is so far no explicit demonstration of such rapid learning with schema reuse as far as we know.

Structured relational representations have been proposed as a common mechanism in the hippocampus that reconciles spatial and non-spatial tasks and memory into the original cognitive mapping view \citep{Eichenbaum2014-dt, Stachenfeld2017-vr}. Several recent studies model cognitive maps as higher order latent graph structures and show generalization to disparate functions implicated in the hippocampus \citep{George2021-qt, raju2022space,Sharma2021-lg,Whittington2020-nu,Whittington2021-ii}. We take one such model of cognitive maps, the clone-structured cognitive graph (CSCG)\citep{George2021-qt}, and extend it to provide a concrete computational model of abstractions using graph schemas. Desiderata for our model is a unified system that facilitates learning new environments by using schemas, handles perceptual aliasing \citep{whitehead1991learning}, and generalizes via map induction \citep{Sharma2021-lg} and schema composition.

Our setup is an agent navigating in an environment modeled as a directed graph. The agent observes the emissions at its current node and traverses to new nodes via edges with labeled actions. Multiple nodes may emit the same observations (i.e., they are aliased \citep{Lajoie2018-hs,whitehead1991learning}), so the agent cannot observe the state directly. This can be considered as a discrete time partially observable Markov decision process (POMDP). 
As the agent navigates a new environment,
our goal is to learn the underlying latent graph (i.e., map induction \cite{Sharma2021-lg}) and to do so rapidly by reusing previously learnt graph topologies or graph schemas.
We choose the CSCG model to construct graph schemas as it has been shown to learn higher-order graphs in highly aliased settings using a smooth, probabilistic, parameterization of the graph learning problem \citep{George2021-qt}. We extend this model to describe how learned graphs can be reused as schemas for transfer learning, quick inference, and planning for behavior in new situations by rapidly learning observation bindings and discovering the best schema online.

\section{Related Work}
Research on rapid transfer to new tasks in novel environments has focused on different aspects, from exploration to modeling and planning. Some of the recent work has been predominantly done in a reinforcement learning (RL) framework, and different RL approaches focus on one or more of these aspects.

\textbf{Meta-RL} Model-free meta-RL approaches focus on exploration policy generalization to new tasks and environmental variations without explicit model learning. Some show generalization to new tasks in a known environment, but not to new environments nor do they handle aliasing \citep{rakelly2019efficient,Wang2016-jr}.
Recurrent model-free RL has shown some generalization to environmental variations and in the POMDP settings \citep{ni2021recurrent}. These variations are of parameters for generating the environment, and the goal is to be able learn an exploration policy that generalizes to these parametric variations. These methods do not re-use explicit knowledge of past environments to model new environments nor do they handle composition of known environments \citep{packer2018assessing}. More recent works added episodic memory with attention heads to selectively attend and reuse stored memories to rapidly adapt to tasks in new environments in few-shot settings (e.g. episodic planning network)\citep{lampinen2021towards,Ritter2020-hu}. These models match optimal planning only after training for billions of steps. Further, none of these approaches handle aliasing, nor explicitly build models of the environment and plan over them. \cite{gupta2017cognitive} show that using explicit model building (mapping) and navigation via planning in spatial environments can handle partial observability and outperform methods without this ability, but this study does not reuse those models to rapidly learn novel environments.

\textbf{Model based RL} Model based RL works learn an explicit model of the environment \citep{gregor2019shaping}, some even under a POMDP setup \citep{igl2018deep}, and can transfer to new tasks within that environment, but cannot transfer to novel environments. Some studies use a version of sub-goal or landmark discovery and reuse to generalize to novel tasks \citep{eysenbach2019search,kipf2019compile,zhang2021world}. However, these approaches do not address partial observability, and more importantly the generalization is restricted to the same environments. Semi-parametric topological memory is a method that extends landmark based navigation to generalize to novel environments \citep{savinov2018semi}. 
This method trains a network to estimate connectivity between observations at different locations and uses this to build a graph of observations in a novel environment. This method does generalize to novel environments, but its performance degrades significantly with increasing aliasing. Further, a human-generated exploration path was used for constructing the graph.

Our work on schema matching and reuse is related to finding correspondence between graphs in different contexts. The first neural network approach to structure mapping was proposed in \cite{Crouse2021-bj}. But this approach is restricted to the matching problem and has no mechanism to resolve or learn new structures or to plan with partially matching schemas.
Another line of related work focused on solving simplified relational tasks inspired by Raven's Progressive Matrices \citep{kerg2022inductive,webb2021emergent}. The main idea is to separate abstract relations from sensory observations during training, and learn the observation mapping to solve new tasks with the same relations but novel mapping, but in a deterministic and simplified setting.

The main contribution of this paper is a unified system that (i) facilitates rapid transfer learning of new environments using schemas via map-induction and composition, (ii) handles perceptual aliasing, (iii) builds explicit latent graph models of environments directly from actions and observations, (iv) supports planning and inference, and (v) and is inspired and grounded in cognitive and neurosciences.
Ours is the first model to combine all these aspects in a single system.

\begin{figure}[ht!]
\begin{center}
\includegraphics[width=0.95\linewidth]{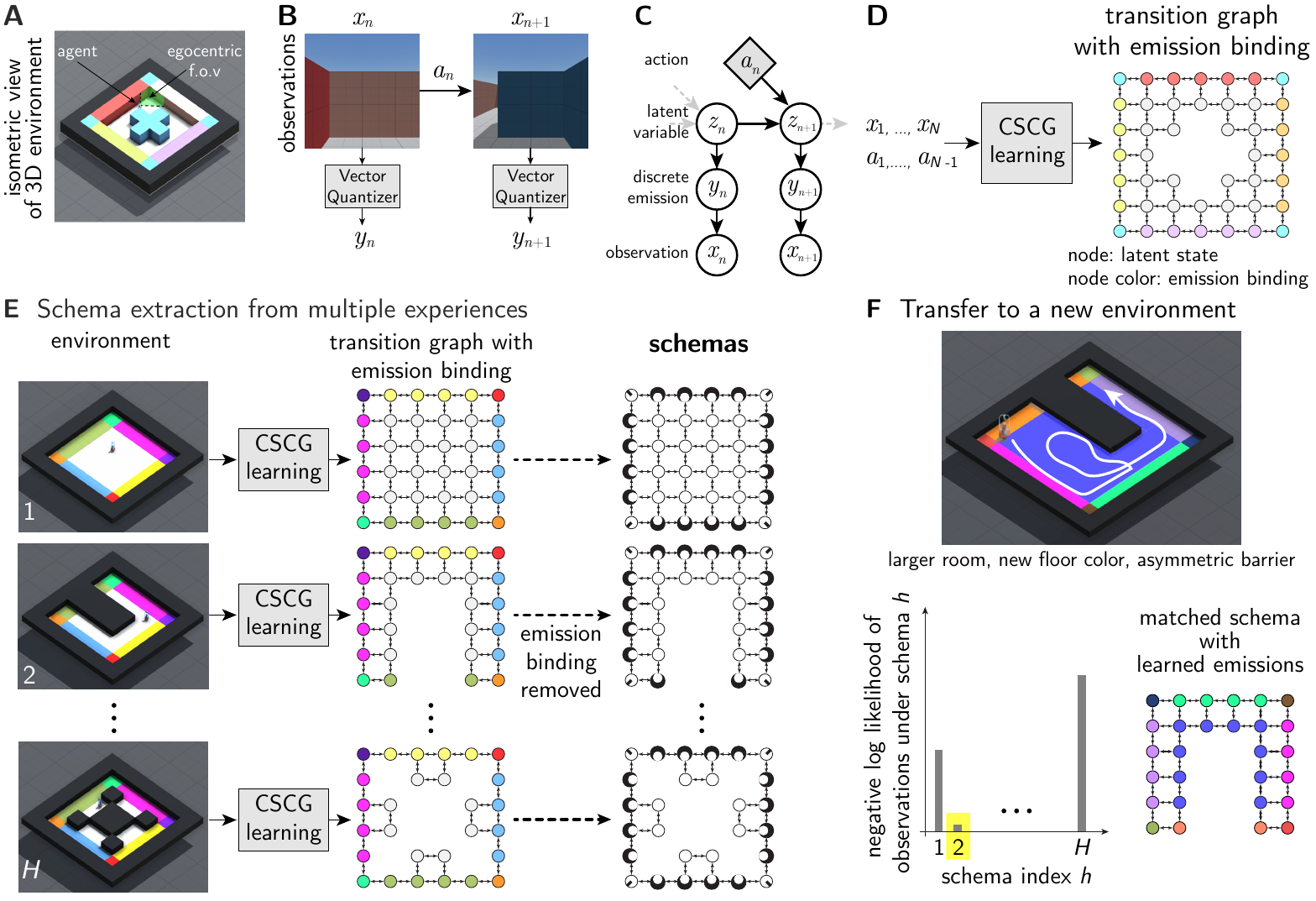}
\end{center}
\caption{Overview of using graph schemas in simulated 3D environments. {\bf A}. An example 3D environment. {\bf B}. The agent navigates the environment with egocentric actions and gets observations as RGB images. The images are passed through a vector quantizer to obtain cluster indices, which are used as observations for training a CSCG. {\bf C}. Graphical model of continuous extension of CSCG with action conditioning. {\bf D}. Given a sequence of actions and observations, a CSCG learns the transition graph of the environment with the corresponding emission binding. We use simplified versions of the transition graphs in this schematic. See Fig. \ref{fig:agix_cscg_models} for real graphs. {\bf E}. Schemas are extracted from learned models with transitions and emissions by unbinding the specific emissions (represented here as floor colors in the environment, and node colors in the corresponding graph), but retaining the clone structure (shown as node patterns). {\bf F}. In a novel environment with size, color, and structure variations, the agent navigates and finds the schema that best explains its observations by learning new emission bindings.}
\label{fig:schematic}
\end{figure}

\section{Methods}
\subsection{Problem Setup}
Consider an agent navigating in a directed graph $G$.
When the agent visits a node in the graph, the node emits an observation. However, multiple nodes may emit the same observation (i.e., they are aliased), so the observation is not enough to disambiguate where in the graph the agent is located. Additionally, actions do not have deterministic results---executing the same action at the same node might result in the agent navigating to different nodes. The outgoing edges from a node are labeled with the action that would traverse them, and with the probability of traversing them under that action. The sum of the probabilities of all outgoing edges from a node with the same action label sum up to 1. We use the graph $G$ to model the agent's environment.

When the agent performs a sequence of actions $a_1,\ldots,a_{N}$ (with discrete $a_n\in\{1,\ldots N_{\rm actions}\}$), it receives a sequence of observations $x_1,\ldots,x_N$ (with each $x_n\in\mathbb{R}^d$ or $\{1, \ldots, N_{\rm obs}\}$ for continuous and discrete observations, respectively). The goal of learning is to recover the topology of the environment $G$ from sequences of actions and observations. The goal of transfer is to reuse the topology of a previously learned graph to model a new environment by relabeling the nodes with new observations. In both cases, the learned graphs can be exploited for tasks such as goal-directed navigation.

\subsection{Model}
\label{subsection:learning_cscg}
Clone-structured cognitive graphs (CSCGs) were introduced in \cite{Dedieu2019-li,George2021-qt} to recover (an approximation of) the graph $G$ from sequences of action-observation pairs. To do this, they use categorical hidden variables $z_1,\ldots,z_N$ to model the node of the graph that the agent is at in each time step. With this, it is possible to formulate a graphical model for the sequence of observations given the actions. Here, we use the conditional version of their model and extend it to continuous observations:
\begin{equation}
p(\boldsymbol{x}| \boldsymbol{a}) = \sum_{\boldsymbol{z}}\sum_{\boldsymbol{y}} P(z_1)\prod_{n=2}^N P(z_n| z_{n-1}, a_{n-1})
    \prod_{n=1}^N P(y_n | z_n)p(x_n | y_n)\label{eqn:ll_schema},
\end{equation}
with $p(x_n | y_n) = \mathcal{N}(x_n |\mu_{y_n}, \sigma^2I)$.
We use the following shorthand for a sequence of actions: $\boldsymbol{a}\equiv\{a_1\ldots,a_N\}$ ($\boldsymbol{x}, \boldsymbol{y}, \boldsymbol{z}$ are similarly defined). The transitions are fully parameterized through an action-conditional transition tensor $T$ with elements  $T_{ijk}= P(z_n=k| z_{n-1}=j, a_{n-1}=i)$. 

To extend CSCG to continuous observations, we introduced a new variable $y_n$ between the hidden state $z_n$ and the observation $x_n$. In this formulation, the observation model is parameterized as an isotropic Gaussian with variance $\sigma^2$ and mean $\mu_{y_n}$, which is the centroid associated to a discrete emission $y_n$. The emission model is parameterized by an emission matrix $E$ with elements $E_{ij}= P(y_n=j| z_n=i)$. In a CSCG, by design, multiple hidden states are forced to share the same emission: if states $i$ and $j$ are \emph{clones} of the same emission, then $p(y|z_i)=p(y|z_j)$. Also, the emissions are deterministic: each row in $E$ has a single entry set to 1, with the remaining elements set to zero. Cloned rows (corresponding to the same emission) will be identical. The idea is that different latent states can produce similar visible observations, and the latent state can only be disambiguated through temporal context. We define $C(x_n)$ as the set of clones of the emission $x_n$.

In a simple case, if we have a finite number of potential observations, we can allocate a centroid to each of them and set $\sigma^2\to 0$. When the observations are discrete, the observation model becomes deterministic and we can set $y=x$. This exactly recovers the discrete CSCG from \cite{George2021-qt}. To learn the discrete CSCG, the clone structure is decided a priori by choosing how many clones to assign to each observation (e.g., a constant number). This fully defines $E$. Then, learning a graph $G$ that describes a new environment only involves learning $T$, i.e.~maximizing $P(\boldsymbol{x}| \boldsymbol{a};T, E)$ w.r.t.~$T$. This can be done via expectation-maximization (EM), see Suppl. \S\ref{Appendix:learn_em_T}. The procedure is analogous to HMM learning, but computationally more efficient by leveraging the sparsity pattern of the emission matrix.

For continuous observations, see Suppl. \S\ref{Appendix:continuous_cscg} for a description of how to learn the centroids $\mu_y$ and allocate hidden states to them (i.e., assigning the clone structure).

\subsection{Schemas for transfer learning}
\label{subsection:schem_reuse}

We call a learned model with fixed 3-tuple $S_g\equiv\{T, C, E\}$ a \textit{grounded schema}. Transfer learning between different environments that share the same graph topology but with different emissions associated to its nodes is possible by reusing the previously learned transitions $T$. We can maximize $p(\boldsymbol{x}| \boldsymbol{a};T, \boldsymbol{\mu})$ w.r.t.~$\boldsymbol{\mu}$ (or $p(\boldsymbol{x}| \boldsymbol{a};T, E)$ w.r.t.~$E$ in the discrete case), while keeping $T$ fixed to its known value from the previous environment (see Suppl. \S\ref{Appendix:learn_em_E} and \S\ref{Appendix:continuous_cscg_transfer} for discrete and continuous observations respectively). We call this $T$ without an associated clone structure $C$ and the emission matrix $E$ as an \textit{ungrounded schema} or simply a \textit{schema}.

Further, if we know that the new environment also preserves the clone structure $C$ (when two nodes were clones in the original environment, they are also clones in the new environment), then we can further restrict the observation model to respect this constraint during learning and reuse the tuple $S_u\equiv\{T, C\}$ and learn $\boldsymbol{\mu}$ or $E$. 
We call this tuple $S_u$ an \textit{ungrounded schema with clone structure} (Fig.~\ref{fig:schematic}E).

For example, in the room navigation setting, a schema models the agent's location and head-directions in a room and how actions move the agent, as well as the knowledge that the floor or door can look the same at multiple locations in the room. Using EM, we show how schemas allow for rapid model learning in new environments that have matching topologies and emission structures with fast binding (Fig.~\ref{fig:schematic}F). Inference can be performed with the matched schemas to actively plan and pursue goals. We can also detect transitions to another known schema, or to unknown territory by comparing likelihoods of observations under different schemas.

Schemas can also be used as building blocks to rapidly learn novel environments that are composed of matching topologies. This comprises learning transitions and emissions but reusing known schemas where they fit (See Suppl. \S\ref{Appendix:schema_stitching} and Suppl. Alg.{2}).

\section{Results}

We show results from two sets of experiments: (i) standard benchmarks for evaluating rapid adaptation, where the environments do not have perceptual aliasing, and (ii) more challenging setup of environments with extensive perceptual aliasing.


\subsection{Rapid adaptation and task solving in novel environments}
We first evaluate our model on two benchmarks proposed by \cite{Ritter2020-hu} to evaluate rapid adaptation and task solving in new environments:  Memory \& Planning Game (MPG) and One-Shot StreetLearn.

\subsubsection{Memory \& Planning Game}
\begin{figure}[ht!]
\begin{center}
\includegraphics[width=0.95\linewidth]{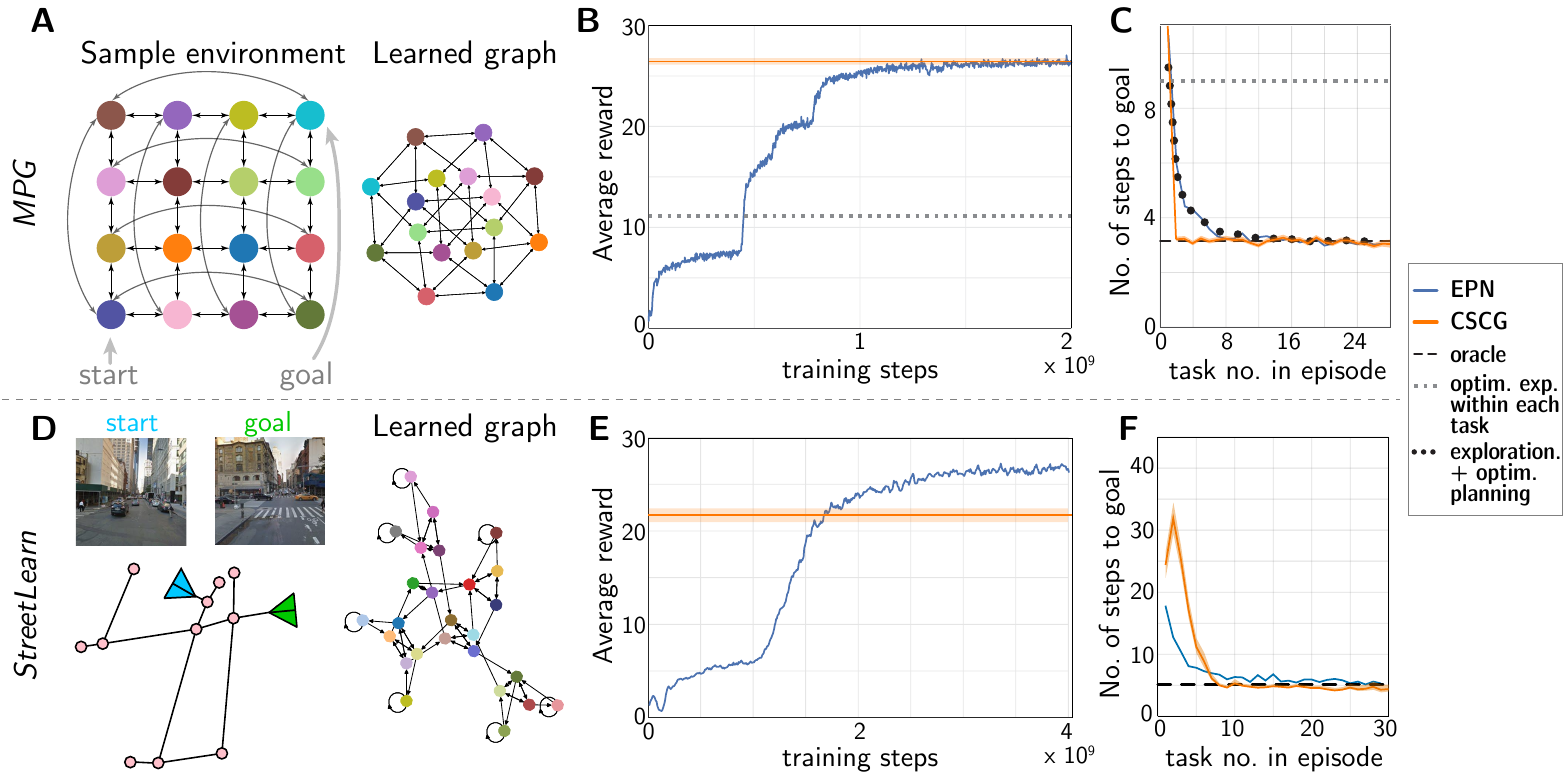}
\end{center}
\caption{Performance on MPG and One-shot StreetLearn benchmarks. Example episodes of MPG (\textbf{A}) and One-Shot StreetLearn (\textbf{D}) along with their respective learned latent graphs. Average rewards in novel test environments (\textbf{B, E}). CSCG performance reaches its peak by 9 episodes in MPG and stays the same thereafter. EPN performance is shown on evaluation at different stages of training. (\textbf{C, F}) Number of steps to get the reward in subsequent tasks in an episode. The CSCG agent explores first without optimizing for reward collection, so our first few tasks take longer to complete but we reach optimal planning after that. Values for optimal exploration and planning are also plotted as dotted lines for comparison. First few tasks in One-Shot StreetLearn take variable numbers of steps as we collect rewards incidentally. EPN baselines and optimal values for both environments are re-plotted using the data from \cite{Ritter2020-hu}. Error bars are 95\% CI of the SEM.}
\label{fig:mpg_streetlearn}
\end{figure}

In the MPG, the agent can navigate on a $4\times 4$ grid, observing symbols, and the task is to collect reward at a specified goal location in that grid (Fig.~\ref{fig:mpg_streetlearn}A). All grid positions have unique symbols and the symbol-position mapping is randomized after each episode, which lasts for 100 steps. See Suppl. \S\ref{Appendix:MPG} for details. This setup lets us evaluate our model on ground truth graph recovery and schema reuse, as the structure is maintained across episodes. The agent needs to explore to collect the observations and bind them rapidly to the schema to maximize rewards with optimal navigation.

\textbf{CSCG schema learns the graph structure in few episodes.} The CSCG schema agent first explores the grid randomly collecting observations for a few episodes. After each episode, we learn a CSCG model that best explains the experiences across all episodes so far observed. We reuse the same schema ($T$) across all episodes and learn a new binding (emission matrix $E$) per episode. It takes only 9 episodes (900 steps) to learn a perfect schema of this $4\times 4$ grid environment. In subsequent episodes, we rapidly learn the schema bindings and do planning to maximize the rewards. We employ two different hard coded exploration policies: random navigation actions and an optimal set of actions to cover the learned transition graph. Average reward per episode $\pm$ standard error of the mean (SEM) after learning the schema is: $17.3 \pm 0.57$ for random and $26.4 \pm 0.17$ for optimal exploration policy, which is comparable to Episodic planning network (EPN) \citep{Ritter2020-hu}. In contrast, EPN takes more than 10 million episodes ($> 10^9$ steps) of training to reach its optimal performance (Fig.~\ref{fig:mpg_streetlearn}B). Planning in our model is optimal in the number of steps to the reward, on par with EPN and the oracle (Fig.~\ref{fig:mpg_streetlearn}C). CSCG performance remains the same since the first reward is collected after exploration and the plans are optimal thereafter. Note that the number of steps to finish the first task is longer in our case ($18 \pm 0.09$ steps) than EPN, but the average reward in an episode is comparable.

\subsubsection{One-Shot StreetLearn}
One-Shot StreetLearn is a challenging variant of the StreetLearn task \citep{mirowski2019streetlearn} with varying connectivity structure across episodes to evaluate rapid task solving ability (Fig.~\ref{fig:mpg_streetlearn}D). In each episode, the agent is placed in a new neighborhood of a city and the task is to navigate to a goal, specified by the goal street view image, and collect the reward. After collecting a reward, the agent is re-spawned in a new location and a new goal is specified. Unlike the MPG, the transition graph changes every episode. We evaluate our model's ability to rapidly learn in an episode and to navigate optimally to the goals to maximize the rewards. Note that there is no schema reuse in this setting: we learn a new model for every episode. This showcases the ability to learn rapidly within a few steps without any prior training and plan efficiently with the learned model.

\textbf{CSCG matches optimal planning in One-Shot StreetLearn}. For the CSCG agent, we follow an explore and exploit strategy with a hard coded exploration policy. During exploration, the agent navigates every action from every observation it encounters while collecting the rewards as it encounters the goals, and uses this experience to learn a CSCG. This is a guided exploration to cover every possible edge in the transition graph. After exploration, the agent plans with the CSCG and collects rewards (See Suppl. \S\ref{Appendix:streetlearn} for details). Average reward $\pm$ SEM over 100 episodes is $21.7 \pm 3.7$, which is lower than EPN (28.7) as our exploration strategy is not optimal (Fig.~\ref{fig:mpg_streetlearn}E). Since we do not consider optimal exploration in this work, we compare the planning performance on the learned model after exploration. Post exploration, our agent takes on average $4.8 \pm 0.03$ steps to reach the goal, which matches the optimal value \citep{Ritter2020-hu} (Fig.~\ref{fig:mpg_streetlearn}F). Note that we do not transfer any learning across episodes in this setting since the graph changes every episode. In cities with re-usable graph structures such as the grid layout in Manhattan, CSCG schemas benefit from the reuse. We evaluate this schema reuse in detail in much harder settings in the following experiments on navigating in rooms with extensively aliased observations.

\begin{figure}[ht!]
\begin{center}
\includegraphics[width=1\linewidth]{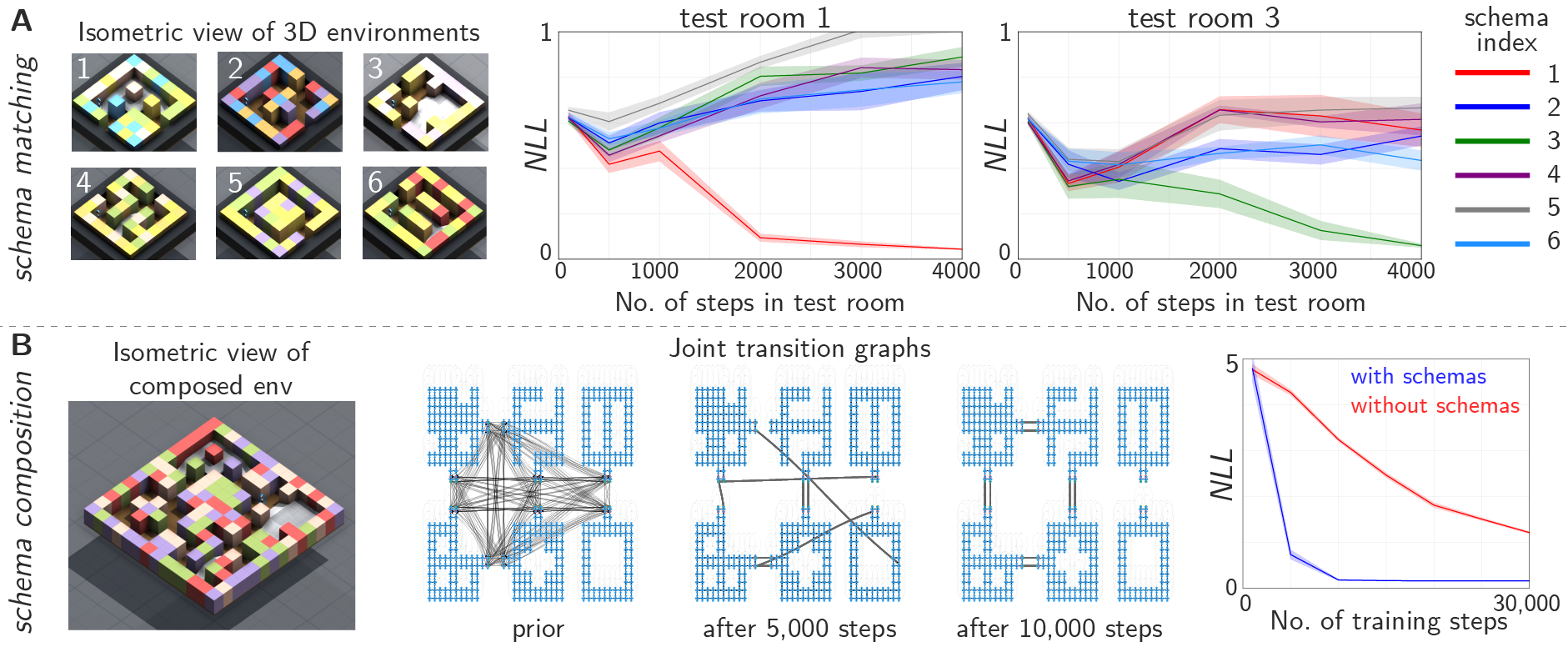}
\end{center}
\caption{Rapid learning using schemas in novel 3D environments. {\bf A}. As the agent walks in a test room, we learn the emission matrix and measure the negative log likelihood (NLL) of observation sequences under different schemas. The schema with the least NLL (conversely the highest likelihood) is considered the best match. The left panel shows the isometric view of the the training rooms used for learning the schemas. The middle and right panels show schema matching for two example test rooms. The test rooms differ in color and lighting conditions from the training rooms.
{\bf B}. Schema stitching in a novel 3D environment (left) composed of four out of the six rooms. Transition graphs show the prior model, with potential connections between all six schemas, and the learned models after 5,000 and 10,000 steps. (Right) NLL of test data for models learned at different walk lengths, for both the model stitched from known schemas and the model learned from scratch.  Error bars are SEM, computed across 5 repetitions.}

\label{fig:schema_stitching}
\end{figure}
%

\subsection{Schema matching and transfer learning in highly aliased environments}
\label{subsection:schema_matching}

In the next set of experiments, we evaluate schema matching and transfer learning in novel environments that vary in observations and in some cases size from training environments. For these experiments, we use a more challenging setting than the benchmarks above, with larger 2D and 3D simulated environments and extensive aliasing \citep{Beattie2016-rk}. Note that even though we use spatial navigation as our setup, we do not use any assumptions about space (Euclidean or otherwise) and model it as a graph navigation problem. For both 2D and 3D environments, we use a simple model of the agent's observations that does not rely on 3D perception, so our results are agnostic to the specific 3D spatial setup.  

We start with a set of environments of different shapes and topologies with extensive aliasing. Analogous to a large empty arena, observations in the interior of these environments are perceptually aliased \cite{Lajoie2018-hs,whitehead1991learning}, see Fig.~\ref{fig:schematic}D. Aliasing occurs by construction in the 2D case, and as a result of clustering in the 3D case. We first learn the schemas in training rooms (Fig.~\ref{fig:schematic}D), as described in section \ref{subsection:schem_reuse} . We evaluate schema matching on test rooms with similar layouts but with novel observations and size variations. In a novel test room, the agent takes a random walk and evaluates the likelihood of the observation sequences given the actions executed under different learned schemas (Fig.~\ref{fig:schematic}E). Note that this requires first learning the new emission matrix from the data collected during this random walk and computing the likelihood for each schema. The schema with the best likelihood is considered the matching schema. We evaluate the likelihoods of different schemas at multiple intervals during the random walks. See Suppl. \S\ref{Appendix:room_schema_recognition} for details and parameters used in the following experiments.

\textbf{CSCG schemas rapidly learn matching bindings in novel environments}. To demonstrate this, we used 3D environments \citep{Beattie2016-rk} with 6 different layouts. The agent can navigate with 3 discrete egocentric actions (move forward, turn left, turn right) and the observations are RGB images corresponding to the agent's view (Fig.~\ref{fig:schematic}A, B). The observation space in this setup is large and complex and demonstrates the model's applicability to such use cases. We follow the procedure for continuous observations described in $\S$\ref{subsection:learning_cscg} and Suppl. \S\ref{Appendix:continuous_cscg} to learn models of the training environments (see Fig. \ref{fig:agix_cscg_models} for the learned graphs after training using random walks). We evaluated rapid learning on test environments with the same layout but with different colored walls, floor, and environment lighting, which corresponds to entirely new RGB observations, for an agent navigating in those rooms using the procedure described in Suppl. \S\ref{Appendix:continuous_cscg_transfer}. Fig.~\ref{fig:schema_stitching}A shows successful rapid matching of correct schemas in test rooms as evaluated by negative log likelihoods (See Suppl. Fig. \ref{fig:agix_schema_recognition} for full results). The correct schema was identified for all six test environments, usually within 1,000 steps in the environment and in all cases within 2,000 steps, compared to about 50,000 steps used to learn without schema reuse.

To test the transfer to size variations, we used 2D rooms of five different layouts (cylinder, rectangle, square with hole, torus, and U-shape) each in 3 different sizes (small, medium, large) (Fig. \ref{fig:all_shape_rooms}).
We learn schemas for the medium-sized versions of these rooms using a random walk of $50,000$ steps. In the test rooms with novel observation mappings, the agent takes a random walk while we learn the new emissions and evaluate the likelihoods of these observations under all schemas every 5 steps. 
See Appendix \ref{Appendix:schema_matching_2D_rooms} for details. Fig.~\ref{fig:all_shape_rooms} shows the negative log likelihoods of all test room under all schemas. By reusing clone structure, we are able to correctly match the schema in all cases by 95 steps demonstrating a rapid matching and adaptation to novel environments with size and observation variations (Appendix Fig.~\ref{fig:schema_matching_no_clone_structure} shows results without using clone structure). We reproduced these results using ten digit exemplars of binarized MNIST dataset as our room layouts as they provide an interesting variety of shapes and topologies that are not designed by us (see Suppl. \S\ref{Appendix:mnist_experiments} for results).

Schema matching also works in environments composed of multiple schemas. We demonstrate this in novel test environments that are composed of pairs of MNIST digit rooms, and show online schema matching of individual digits. See Suppl. \S\ref{Appendix:schema_stitching} for details and results.

\subsection{Rapid learning of novel environments with compositionality of graph schemas}
\label{schema_stitching}
Schemas can also be used to efficiently learn and navigate in larger environments composed of known schemas. We do this by matching schemas and learning the transition structure between them. Fig. \ref{fig:schema_stitching}B shows an example 3D simulated environment, composed of four smaller 3D rooms. The agent walks in this environment and learns the composed model (both joint transitions and emissions) in far fewer steps than needed to learn without using schemas. Fig. \ref{fig:schema_stitching}B also shows the prior model with all potential connections and the learned model at two different walk lengths. Model quality is measured by the negative log likelihood over $10,000$-step test walks, for models trained after walks of different lengths with and without using schemas. Using schemas, we were able to learn a perfect model of the environment in less than $10,000$ steps, whereas learning from scratch was significantly worse even after $30,000$ steps. See Suppl. \S\ref{Appendix:schema_stitching} for the learning algorithm and experimental details. 
This ability to learn by composing and reusing previously learned schemas enables rapid adaptation to novel environments which only gets better with more experience.

\begin{figure}[ht!]
\begin{center}
\includegraphics[width=1\linewidth]{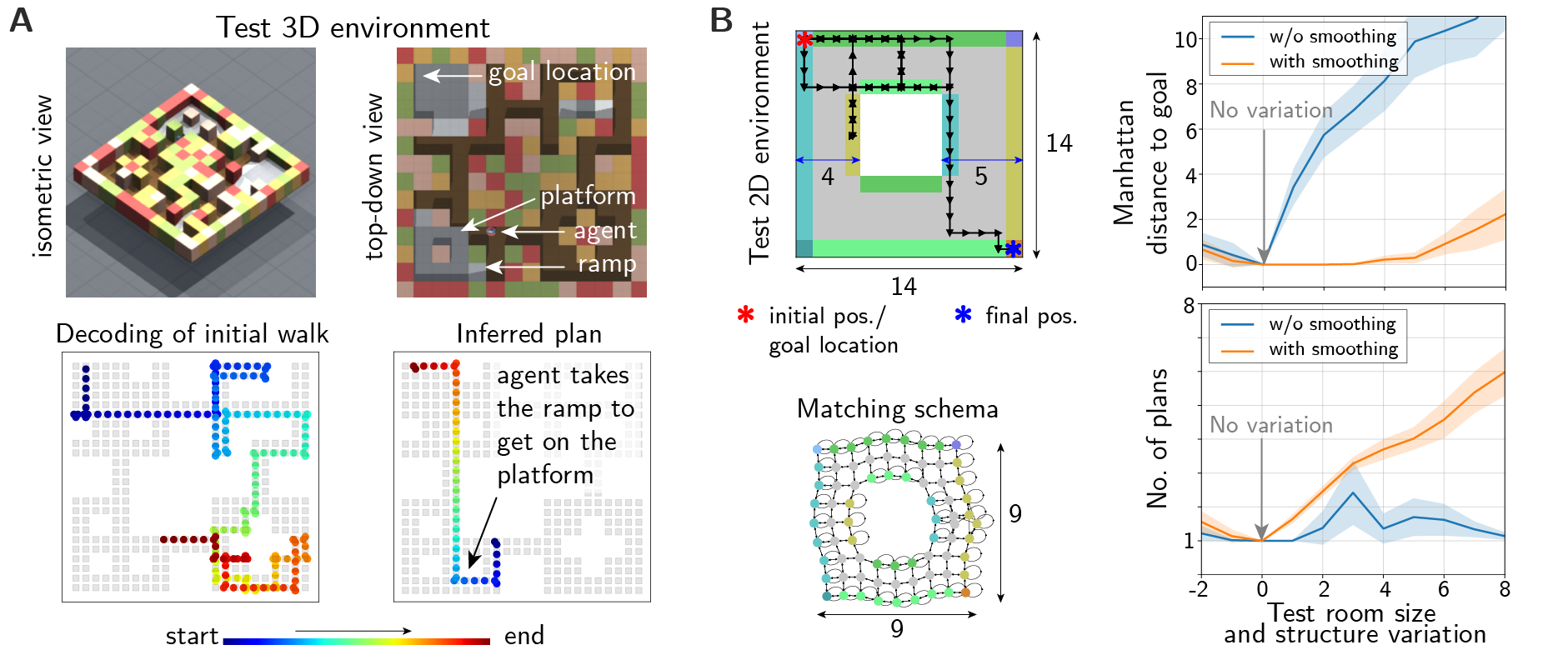}
\end{center}
\caption{Schemas enable rapid and robust planning in new environments. {\bf A}. Given partial experience in a novel environment (here, a differently colored variation of the four-room environment from Fig. \ref{fig:schema_stitching}), the joint transition model and the agent's location within it can be identified and used to rapidly navigate around obstacles and find the shortest path to the goal (here, to return to the starting position). {\bf B}. (Left) An example test room and its matching schema. Note the mismatch in the size and the off-center hole in the test room. (Right). Navigation performance as measured by the distance from the true goal and the number of plans/re-plans the agent makes, as a function of size and structure variation between test room and the matching schema. Error bars are 95\% CI of the SEM.}
\label{fig:shortcut_planning_results}
\end{figure}

\subsection{Rapid planning and navigation in novel environments}
Rapid schema matching and binding enables planning in novel environments with limited experience. We first demonstrate this capacity in a novel variation of the four-room 3D environment that was introduced in Fig. \ref{fig:shortcut_planning_results}A. In this demonstration, an agent first walks in the test environment, and is then tasked with finding the shortest route back to its starting location. We used a manually specified initial walk in order to cover a good portion of the environment in a minimal number of steps. In this case, the schema -- the composite schema that was learned previously for the four-room combination (Fig \ref{fig:schema_stitching}) -- is known a priori. The agent uses the observations from the initial walk to learn new emission bindings and decode its current and goal positions within the model. The agent then plans the shortest path to the goal using this model (Fig. \ref{fig:shortcut_planning_results}B). Note that the planned route passes through a yet unvisited part of the environment.

While executing a plan, if the agent estimates that it has not reached the goal after accounting for the new observations, it is possible that there is a schema mismatch or the estimated emission matrix is inaccurate. When this occurs, the agent can initiate re-planning after updating the model using new experiences gained while executing the plan. This process is iterated until the agent believes that it has reached its goal after decoding observations from the initial random walk and all subsequent re-planning steps. In an experiment designed to test this capacity, we systematically evaluate this robustness to schema mismatch in 2D environments in terms of Manhattan distance from the goal location, and the number of re-plans required. 
Fig.~\ref{fig:shortcut_planning_results}B shows results for one example schema. We can successfully navigate even with size and structure variations and the performance degrades gracefully as the variation between the schema and the test room increases. The number of planning attempts required to reach the goal also increases smoothly. The diagonal smoothing term that adds self-transition probabilities (see Suppl. \S\ref{Appendix:shortcut_finding}) is critical for generalization to size variations. Without this smoothing, the agent never reaches the goal in larger size variations of the test rooms as shown in Fig ~\ref{fig:shortcut_planning_results}B (See Suppl. \S\ref{Appendix:shortcut_finding} for results in another room layout).

\section{Discussion and Future Work}

Learning abstractions that can rapidly bind to observations from environments that share the same underlying structure is the hypothesized mechanism for transfer learning in humans and animals \citep{Kumaran2016-gz,Tse2007-tg,Zhou2021-ja}. We have proposed a concrete computational model of abstractions and rapid binding using graph schemas that learn higher-order structures from aliased observation sequences, and uses a slot-binding mechanism for transferring those schemas to rapidly learn models of new environments. 
CSCG schemas learned graph structure in far fewer episodes than a deep RL agent and matched optimal planning in MPG and One-Shot StreetLearn tasks. In highly aliased environments, CSCG schemas found matching schema bindings in novel rooms of different sizes. In composed rooms, we were able to match the correct schemas as the agent moved across rooms corresponding to different schemas. We showed successful planning to goals in rooms with shape and size variations from the matching schema by re-planning and updating the model while walking to the goal. More importantly, known schemas can be composed to rapidly learn novel environments and new larger schemas. This ability bootstraps on itself and only gets better with more experience. There are many clear directions for potential future works to build on top of our current work, and we list some of them below.

\textbf{Schema learning from experiences.} We learned schemas independently and explicitly in this work, but in the real world, it might not be feasible to have access to differentiated experiences belonging to distinct schemas. Learning reusable schemas from a continuous stream of experiences \citep{farzanfar2023cognitive} could be an interesting future work.

\textbf{Schemas vs memories.} We discard previously learned emissions from past experiences and learn a new binding. However, in some cases, previously learned emissions are directly applicable and therefore keeping those in addition could enable even faster zero-shot adaptation when there is a match. This can be thought as keeping specific memories versus using abstract schemas.

\textbf{Schema maintenance.} Our schemas in this work are fixed. However, it is possible to update the schemas with new experiences. In fact, children initially tend to perceive and remember experiences that fit in their existing schemas and develop the flexibility later \citep{piaget1952origins}. Similarly, we could update schemas based on new experiences, and even make the schemas themselves flexible to encapsulate related abstractions but still constrained by rules to allow consistent inference. 

\textbf{Active exploration.} We used either random or known optimal exploration policies to learn and bind schemas. But the schemas provide action-conditioned beliefs on future observations, and by choosing actions that could optimally disambiguate different schemas and seek to learn connections between them, we could potentially do much better than random exploration. Similarly, instead of random exploration to learn a new environment and schema, we could direct the exploration policy by composing known schemas \citep{Sharma2021-lg}, and even actively learn them while exploring.

\bibliography{main}

\begin{thebibliography}{44}
\providecommand{\natexlab}[1]{#1}
\providecommand{\url}[1]{\texttt{#1}}
\expandafter\ifx\csname urlstyle\endcsname\relax
  \providecommand{\doi}[1]{doi: #1}\else
  \providecommand{\doi}{doi: \begingroup \urlstyle{rm}\Url}\fi

\bibitem[Baraduc et~al.(2019)Baraduc, Duhamel, and Wirth]{Baraduc2019-ro}
P.~Baraduc, J.-R. Duhamel, and S.~Wirth.
\newblock Schema cells in the macaque hippocampus.
\newblock \emph{Science}, 363\penalty0 (6427):\penalty0 635--639, Feb. 2019.

\bibitem[Beattie et~al.(2016)Beattie, Leibo, Teplyashin, Ward, Wainwright,
  Küttler, Lefrancq, Green, Valdés, Sadik, Schrittwieser, Anderson, York,
  Cant, Cain, Bolton, Gaffney, King, Hassabis, Legg, and
  Petersen]{Beattie2016-rk}
C.~Beattie, J.~Z. Leibo, D.~Teplyashin, T.~Ward, M.~Wainwright, H.~Küttler,
  A.~Lefrancq, S.~Green, V.~Valdés, A.~Sadik, J.~Schrittwieser, K.~Anderson,
  S.~York, M.~Cant, A.~Cain, A.~Bolton, S.~Gaffney, H.~King, D.~Hassabis,
  S.~Legg, and S.~Petersen.
\newblock Deepmind lab, 2016.
\newblock URL \url{https://arxiv.org/abs/1612.03801}.

\bibitem[Cormack and Horspool(1987)]{cormack1987data}
G.~V. Cormack and R.~N.~S. Horspool.
\newblock {Data compression using dynamic {M}arkov modelling}.
\newblock \emph{The Computer Journal}, 30\penalty0 (6):\penalty0 541--550,
  1987.

\bibitem[Crouse et~al.(2021)Crouse, Nakos, Abdelaziz, and
  Forbus]{Crouse2021-bj}
M.~Crouse, C.~Nakos, I.~Abdelaziz, and K.~Forbus.
\newblock Neural analogical matching.
\newblock \emph{AAAI}, 35\penalty0 (1):\penalty0 809--817, May 2021.

\bibitem[Dedieu et~al.(2019)Dedieu, Gothoskar, Swingle, Lehrach,
  Lázaro-Gredilla, and George]{Dedieu2019-li}
A.~Dedieu, N.~Gothoskar, S.~Swingle, W.~Lehrach, M.~Lázaro-Gredilla, and
  D.~George.
\newblock Learning higher-order sequential structure with cloned hmms, 2019.
\newblock URL \url{https://arxiv.org/abs/1905.00507}.

\bibitem[Eichenbaum and Cohen(2014)]{Eichenbaum2014-dt}
H.~Eichenbaum and N.~J. Cohen.
\newblock Can we reconcile the declarative memory and spatial navigation views
  on hippocampal function?
\newblock \emph{Neuron}, 83\penalty0 (4):\penalty0 764--770, Aug. 2014.

\bibitem[Eysenbach et~al.(2019)Eysenbach, Salakhutdinov, and
  Levine]{eysenbach2019search}
B.~Eysenbach, R.~R. Salakhutdinov, and S.~Levine.
\newblock Search on the replay buffer: Bridging planning and reinforcement
  learning.
\newblock \emph{Advances in Neural Information Processing Systems}, 32, 2019.

\bibitem[Farzanfar et~al.(2023)Farzanfar, Spiers, Moscovitch, and
  Rosenbaum]{farzanfar2023cognitive}
D.~Farzanfar, H.~J. Spiers, M.~Moscovitch, and R.~S. Rosenbaum.
\newblock From cognitive maps to spatial schemas.
\newblock \emph{Nature Reviews Neuroscience}, 24\penalty0 (2):\penalty0 63--79,
  2023.

\bibitem[George et~al.(2021)George, Rikhye, Gothoskar, Guntupalli, Dedieu, and
  L{\'a}zaro-Gredilla]{George2021-qt}
D.~George, R.~V. Rikhye, N.~Gothoskar, J.~S. Guntupalli, A.~Dedieu, and
  M.~L{\'a}zaro-Gredilla.
\newblock Clone-structured graph representations enable flexible learning and
  vicarious evaluation of cognitive maps.
\newblock \emph{Nat. Commun.}, 12\penalty0 (1):\penalty0 2392, Apr. 2021.

\bibitem[Gilboa and Marlatte(2017)]{Gilboa2017-bk}
A.~Gilboa and H.~Marlatte.
\newblock Neurobiology of schemas and {Schema-Mediated} memory.
\newblock \emph{Trends Cogn. Sci.}, 21\penalty0 (8):\penalty0 618--631, Aug.
  2017.

\bibitem[Gregor et~al.(2019)Gregor, Jimenez~Rezende, Besse, Wu, Merzic, and
  van~den Oord]{gregor2019shaping}
K.~Gregor, D.~Jimenez~Rezende, F.~Besse, Y.~Wu, H.~Merzic, and A.~van~den Oord.
\newblock Shaping belief states with generative environment models for rl.
\newblock \emph{Advances in Neural Information Processing Systems}, 32, 2019.

\bibitem[Gupta et~al.(2017)Gupta, Davidson, Levine, Sukthankar, and
  Malik]{gupta2017cognitive}
S.~Gupta, J.~Davidson, S.~Levine, R.~Sukthankar, and J.~Malik.
\newblock Cognitive mapping and planning for visual navigation.
\newblock In \emph{Proceedings of the IEEE conference on computer vision and
  pattern recognition}, pages 2616--2625, 2017.

\bibitem[Igl et~al.(2018)Igl, Zintgraf, Le, Wood, and Whiteson]{igl2018deep}
M.~Igl, L.~Zintgraf, T.~A. Le, F.~Wood, and S.~Whiteson.
\newblock Deep variational reinforcement learning for pomdps.
\newblock In \emph{International Conference on Machine Learning}, pages
  2117--2126. PMLR, 2018.

\bibitem[Kerg et~al.(2022)Kerg, Mittal, Rolnick, Bengio, Richards, and
  Lajoie]{kerg2022inductive}
G.~Kerg, S.~Mittal, D.~Rolnick, Y.~Bengio, B.~A. Richards, and G.~Lajoie.
\newblock Inductive biases for relational tasks.
\newblock In \emph{ICLR2022 Workshop on the Elements of Reasoning: Objects,
  Structure and Causality}, 2022.

\bibitem[Kipf et~al.(2019)Kipf, Li, Dai, Zambaldi, Sanchez-Gonzalez,
  Grefenstette, Kohli, and Battaglia]{kipf2019compile}
T.~Kipf, Y.~Li, H.~Dai, V.~Zambaldi, A.~Sanchez-Gonzalez, E.~Grefenstette,
  P.~Kohli, and P.~Battaglia.
\newblock Compile: Compositional imitation learning and execution.
\newblock In \emph{International Conference on Machine Learning}, pages
  3418--3428. PMLR, 2019.

\bibitem[Kumaran et~al.(2016)Kumaran, Hassabis, and McClelland]{Kumaran2016-gz}
D.~Kumaran, D.~Hassabis, and J.~L. McClelland.
\newblock What learning systems do intelligent agents need? complementary
  learning systems theory updated.
\newblock \emph{Trends Cogn. Sci.}, 20\penalty0 (7):\penalty0 512--534, July
  2016.

\bibitem[Lajoie et~al.(2019)Lajoie, Hu, Beltrame, and Carlone]{Lajoie2018-hs}
P.-Y. Lajoie, S.~Hu, G.~Beltrame, and L.~Carlone.
\newblock Modeling perceptual aliasing in slam via discrete–continuous
  graphical models.
\newblock \emph{IEEE Robotics and Automation Letters}, 4\penalty0 (2):\penalty0
  1232--1239, 2019.
\newblock \doi{10.1109/LRA.2019.2894852}.

\bibitem[Lampinen et~al.(2021)Lampinen, Chan, Banino, and
  Hill]{lampinen2021towards}
A.~Lampinen, S.~Chan, A.~Banino, and F.~Hill.
\newblock Towards mental time travel: a hierarchical memory for reinforcement
  learning agents.
\newblock \emph{Advances in Neural Information Processing Systems},
  34:\penalty0 28182--28195, 2021.

\bibitem[Mirowski et~al.(2019)Mirowski, Banki-Horvath, Anderson, Teplyashin,
  Hermann, Malinowski, Grimes, Simonyan, Kavukcuoglu, Zisserman,
  et~al.]{mirowski2019streetlearn}
P.~Mirowski, A.~Banki-Horvath, K.~Anderson, D.~Teplyashin, K.~M. Hermann,
  M.~Malinowski, M.~K. Grimes, K.~Simonyan, K.~Kavukcuoglu, A.~Zisserman,
  et~al.
\newblock The streetlearn environment and dataset.
\newblock \emph{arXiv preprint arXiv:1903.01292}, 2019.

\bibitem[Mitchell(2021)]{Mitchell2021-pc}
M.~Mitchell.
\newblock Abstraction and analogy-making in artificial intelligence.
\newblock \emph{Ann. N. Y. Acad. Sci.}, 1505\penalty0 (1):\penalty0 79--101,
  Dec. 2021.

\bibitem[Ni et~al.(2021)Ni, Eysenbach, and Salakhutdinov]{ni2021recurrent}
T.~Ni, B.~Eysenbach, and R.~Salakhutdinov.
\newblock Recurrent model-free rl can be a strong baseline for many pomdps.
\newblock \emph{arXiv preprint arXiv:2110.05038}, 2021.

\bibitem[Packer et~al.(2018)Packer, Gao, Kos, Kr{\"a}henb{\"u}hl, Koltun, and
  Song]{packer2018assessing}
C.~Packer, K.~Gao, J.~Kos, P.~Kr{\"a}henb{\"u}hl, V.~Koltun, and D.~Song.
\newblock Assessing generalization in deep reinforcement learning.
\newblock \emph{arXiv preprint arXiv:1810.12282}, 2018.

\bibitem[Piaget and Cook(1952)]{piaget1952origins}
J.~Piaget and M.~T. Cook.
\newblock \emph{The origins of intelligence in children.}
\newblock WW Norton \& Co, 1952.

\bibitem[Preston and Eichenbaum(2013)]{Preston2013-pu}
A.~R. Preston and H.~Eichenbaum.
\newblock Interplay of hippocampus and prefrontal cortex in memory.
\newblock \emph{Curr. Biol.}, 23\penalty0 (17):\penalty0 R764--73, Sept. 2013.

\bibitem[Raju et~al.(2022)Raju, Guntupalli, Zhou, Lázaro-Gredilla, and
  George]{raju2022space}
R.~V. Raju, J.~S. Guntupalli, G.~Zhou, M.~Lázaro-Gredilla, and D.~George.
\newblock Space is a latent sequence: Structured sequence learning as a unified
  theory of representation in the hippocampus, 2022.

\bibitem[Rakelly et~al.(2019)Rakelly, Zhou, Finn, Levine, and
  Quillen]{rakelly2019efficient}
K.~Rakelly, A.~Zhou, C.~Finn, S.~Levine, and D.~Quillen.
\newblock Efficient off-policy meta-reinforcement learning via probabilistic
  context variables.
\newblock In \emph{International conference on machine learning}, pages
  5331--5340. PMLR, 2019.

\bibitem[Ritter et~al.(2020)Ritter, Faulkner, Sartran, Santoro, Botvinick, and
  Raposo]{Ritter2020-hu}
S.~Ritter, R.~Faulkner, L.~Sartran, A.~Santoro, M.~Botvinick, and D.~Raposo.
\newblock Rapid task-solving in novel environments, 2020.
\newblock URL \url{https://arxiv.org/abs/2006.03662}.

\bibitem[Samborska et~al.(2022)Samborska, Butler, Walton, Behrens, and
  Akam]{Samborska2022-vn}
V.~Samborska, J.~L. Butler, M.~E. Walton, T.~E.~J. Behrens, and T.~Akam.
\newblock Complementary task representations in hippocampus and prefrontal
  cortex for generalizing the structure of problems.
\newblock \emph{Nat. Neurosci.}, pages 1--13, Sept. 2022.

\bibitem[Savinov et~al.(2018)Savinov, Dosovitskiy, and Koltun]{savinov2018semi}
N.~Savinov, A.~Dosovitskiy, and V.~Koltun.
\newblock Semi-parametric topological memory for navigation.
\newblock \emph{arXiv preprint arXiv:1803.00653}, 2018.

\bibitem[Shanahan and Mitchell(2022)]{Shanahan2022-lp}
M.~Shanahan and M.~Mitchell.
\newblock Abstraction for deep reinforcement learning, 2022.
\newblock URL \url{https://arxiv.org/abs/2202.05839}.

\bibitem[Sharan et~al.(2017)Sharan, Kakade, Liang, and
  Valiant]{sharan2017learning}
V.~Sharan, S.~M. Kakade, P.~S. Liang, and G.~Valiant.
\newblock Learning overcomplete hmms.
\newblock In \emph{Advances in Neural Information Processing Systems}, pages
  940--949, 2017.

\bibitem[Sharma et~al.(2021)Sharma, Curtis, Kryven, Tenenbaum, and
  Fiete]{Sharma2021-lg}
S.~Sharma, A.~Curtis, M.~Kryven, J.~Tenenbaum, and I.~Fiete.
\newblock Map induction: Compositional spatial submap learning for efficient
  exploration in novel environments, 2021.
\newblock URL \url{https://arxiv.org/abs/2110.12301}.

\bibitem[Stachenfeld et~al.(2017)Stachenfeld, Botvinick, and
  Gershman]{Stachenfeld2017-vr}
K.~L. Stachenfeld, M.~M. Botvinick, and S.~J. Gershman.
\newblock The hippocampus as a predictive map.
\newblock \emph{Nat. Neurosci.}, 20, 2017.

\bibitem[Tenenbaum et~al.(2011)Tenenbaum, Kemp, Griffiths, and
  Goodman]{Tenenbaum2011-cl}
J.~B. Tenenbaum, C.~Kemp, T.~L. Griffiths, and N.~D. Goodman.
\newblock How to grow a mind: statistics, structure, and abstraction.
\newblock \emph{Science}, 331\penalty0 (6022):\penalty0 1279--1285, Mar. 2011.

\bibitem[Tse et~al.(2007)Tse, Langston, Kakeyama, Bethus, Spooner, Wood,
  Witter, and Morris]{Tse2007-tg}
D.~Tse, R.~F. Langston, M.~Kakeyama, I.~Bethus, P.~A. Spooner, E.~R. Wood,
  M.~P. Witter, and R.~G.~M. Morris.
\newblock Schemas and memory consolidation.
\newblock \emph{Science}, 316\penalty0 (5821):\penalty0 76--82, Apr. 2007.

\bibitem[Verbeek et~al.(2003)Verbeek, Vlassis, and
  Kr{\"o}se]{verbeek2003efficient}
J.~J. Verbeek, N.~Vlassis, and B.~Kr{\"o}se.
\newblock Efficient greedy learning of gaussian mixture models.
\newblock \emph{Neural computation}, 15\penalty0 (2):\penalty0 469--485, 2003.

\bibitem[Wang et~al.(2016)Wang, Kurth-Nelson, Tirumala, Soyer, Leibo, Munos,
  Blundell, Kumaran, and Botvinick]{Wang2016-jr}
J.~X. Wang, Z.~Kurth-Nelson, D.~Tirumala, H.~Soyer, J.~Z. Leibo, R.~Munos,
  C.~Blundell, D.~Kumaran, and M.~Botvinick.
\newblock Learning to reinforcement learn, 2016.
\newblock URL \url{https://arxiv.org/abs/1611.05763}.

\bibitem[Webb et~al.(2021)Webb, Sinha, and Cohen]{webb2021emergent}
T.~W. Webb, I.~Sinha, and J.~Cohen.
\newblock Emergent symbols through binding in external memory.
\newblock In \emph{International Conference on Learning Representations}, 2021.
\newblock URL \url{https://openreview.net/forum?id=LSFCEb3GYU7}.

\bibitem[Whitehead and Ballard(1991)]{whitehead1991learning}
S.~D. Whitehead and D.~H. Ballard.
\newblock Learning to perceive and act by trial and error.
\newblock \emph{Machine Learning}, 7:\penalty0 45--83, 1991.

\bibitem[Whittington et~al.(2020)Whittington, Muller, Mark, Chen, Barry,
  Burgess, and Behrens]{Whittington2020-nu}
J.~C.~R. Whittington, T.~H. Muller, S.~Mark, G.~Chen, C.~Barry, N.~Burgess, and
  T.~E.~J. Behrens.
\newblock The {Tolman-Eichenbaum} machine: Unifying space and relational memory
  through generalization in the hippocampal formation.
\newblock \emph{Cell}, 183\penalty0 (5):\penalty0 1249--1263.e23, Nov. 2020.

\bibitem[Whittington et~al.(2021)Whittington, Warren, and
  Behrens]{Whittington2021-ii}
J.~C.~R. Whittington, J.~Warren, and T.~E.~J. Behrens.
\newblock Relating transformers to models and neural representations of the
  hippocampal formation, 2021.
\newblock URL \url{https://arxiv.org/abs/2112.04035}.

\bibitem[Wu et~al.(1983)]{wu1983convergence}
C.~J. Wu et~al.
\newblock On the convergence properties of the em algorithm.
\newblock \emph{The Annals of statistics}, 11\penalty0 (1):\penalty0 95--103,
  1983.

\bibitem[Zhang et~al.(2021)Zhang, Yang, and Stadie]{zhang2021world}
L.~Zhang, G.~Yang, and B.~C. Stadie.
\newblock World model as a graph: Learning latent landmarks for planning.
\newblock In \emph{International Conference on Machine Learning}, pages
  12611--12620. PMLR, 2021.

\bibitem[Zhou et~al.(2021)Zhou, Jia, Montesinos-Cartagena, Gardner, Zong, and
  Schoenbaum]{Zhou2021-ja}
J.~Zhou, C.~Jia, M.~Montesinos-Cartagena, M.~P.~H. Gardner, W.~Zong, and
  G.~Schoenbaum.
\newblock Evolving schema representations in orbitofrontal ensembles during
  learning.
\newblock \emph{Nature}, 590\penalty0 (7847):\penalty0 606--611, Feb. 2021.

\end{thebibliography}
\newpage
\appendix
\beginsupplement
\onecolumn
\section*{Supplementary material}

\section{Expectation-Maximization learning of CSCGs}
\label{Appendix:CSCG}
Cloned Hidden Markov Models (HMMs), first introduced in \citep{Dedieu2019-li}, are a sparse restriction of overcomplete HMMs \citep{sharan2017learning} that can overcome many of the training shortcomings of dynamic Markov coding \citep{cormack1987data}. Similar to HMMs, cloned HMMs assume the observed sequence $\boldsymbol{x}\equiv\{x_1\ldots,x_N\}$ is generated from a hidden process $\boldsymbol{z}\equiv\{z_1\ldots,z_N\}$ that obeys the Markovian property,
\begin{equation}
    P(\boldsymbol{x}) = \sum_{\boldsymbol{z}}
    P(z_1) \prod_{n=2}^{N} P(z_{n} | z_{n-1}) \prod_{n=1}^{N}P(x_n|z_n).
\end{equation}
Here $P(z_1)$ is the initial hidden state distribution, $P(z_{n} | z_{n-1})$ is the transition probability from $z_{n-1}$ to $z_{n}$, and $P(x_n | z_n)$ is the probability of emitting $x_n$ from the hidden state $z_n$.

In contrast to HMMs, cloned HMMs assume that each hidden state maps deterministically to a single observation. Further, cloned HMMs allow multiple hidden states to emit the same observation. All the hidden states that emit the same observation are called the {\it clones} of that observation. With this constraint, the joint distribution of the observed data can be expressed as,
\begin{equation}
P(\boldsymbol{x}) =  \sum_{z_1 \in C(x_1)} \ldots \sum_{z_N \in C(x_N)}  P(z_1) 
\prod_{n=2}^{N} P(z_{n}|z_{n-1})
\end{equation}
where $C(x_n)$ corresponds to the hidden states (clones) of the emission $x_n$.

Clone-structured cognitive graphs (CSCGs) build on top of cloned HMMs by augmenting the model with the actions of an agent $\boldsymbol{a}\equiv\{a_1\ldots,a_{N}\}$ \citep{George2021-qt},
\begin{equation}
P(\boldsymbol{x} | \boldsymbol{a}) =  \sum_{z_1 \in C(x_1)} \ldots \sum_{z_N \in C(x_N)}  P(z_1) 
\prod_{n=2}^{N} P(z_{n}|z_{n-1}, a_{n-1})
\end{equation}

\subsection{Learning the transition matrix with emissions fixed}
\label{Appendix:learn_em_T}

Learning a CSCG in a new environment requires optimizing the vector of prior probabilities $\pi$: $\pi_k = P(z_1 = k)$ and the action-augmented transition tensor $T$: $T_{ijk} = P(z_n=k| z_{n-1}=j, a_{n-1}=i)$. The emission matrix $E$, which encodes the particular allocation of clones to observations, is kept fixed throughout learning.

The standard algorithm to train HMMs is the expectation-maximization (EM) algorithm \citep{wu1983convergence}, which in this context is known as the Baum-Welch algorithm.
Learning a CSCG using the Baum-Welch algorithm requires a few simple modifications: the sparsity of the emission matrix can be exploited to only use small blocks of the transition matrix both in the Expectation (E) and Maximization (M) steps.

To this end, we assume the hidden states are indexed such that all the clones of the first emission appear first, all the clones of the second emission appear next, etc. Let $N_{\textrm{obs}}$ and $N_{\textrm{actions}}$ be the total number of emissions and actions, respectively. The transition matrix $T$ can then be broken down into smaller submatrices $T(u,v,w)$, where $v,w \in \{1, \ldots, N_{\textrm{obs}} \}$ and $u \in \{1, \ldots, N_{\textrm{actions}} \} $. The submatrix $T(u,v,w)$ contains the transition probabilities $P(z_{n}| z_{n-1}, a_{n-1}=u)$ for $z_{n-1}\in C(v)$ and $z_{n} \in C(w)$, where $C(v)$ and $C(w)$ correspond to the clones of emissions $v$ and $w$ respectively. 

The standard Baum-Welch equations can then be expressed in a simpler form in the case of a CSCG. The E-step recursively computes the forward and backward probabilities and then updates the posterior probabilities. The M-step updates the transition matrix via row normalization.

\textbf{E-Step:}
\begin{align}
\alpha(1) &= \pi(z_1)  &
\alpha(n)^\top &= \alpha(n-1)^\top T(a_{n-1}, C(x_{n-1}), C(x_{n}))\\
\beta(N) &= 1(z_N)  & \beta(n) &= T(a_n, C(x_n), C(x_{n+1})) \beta(n+1)
\end{align}
\begin{align}
\xi_{uvw}(n) &=  \frac{\alpha(n)\circ T(a_n, v, w) \circ \beta(n+1)^\top}{\alpha(n)^\top T(a_n,v, w) \beta(n+1)}\\
\gamma(n) &=  \frac{\alpha(n) \circ \beta(n)}{\alpha(n)^\top\beta(n)}.
\end{align}

\textbf{M-Step:}
\begin{align} 
\pi(z_1) &= \gamma(1)\\
T(u, v, w) &=  \sum_{n=1}^N \xi_{uvw}(n) \oslash \sum_{u=1}^{N_{\textrm{actions}}}  \sum_{w=1}^{N_{\textrm{obs}}} \sum_{n=1}^N \xi_{uvw}(n)
\end{align}

where $\circ$ and $\oslash$ denote element-wise product and division, respectively (with broadcasting where needed). All vectors in the E and M-steps are $N_{\textrm{cpe}}\times1$ column vectors, where $N_{\textrm{cpe}}$ is the number of clones per emission. We use a constant number of clones per emission for simplicity of description, but the number of clones can be selected independently per emission. 

Importantly, CSCGs exploit the sparsity pattern in the emission matrix when performing training updates and inference, and achieve significant computational savings when compared with HMMs \citep{George2021-qt}.

\subsection{Learning the emission matrix with transitions fixed}
\label{Appendix:learn_em_E}
With a CSCG, transfer learning between different environments can be accomplished by keeping its transition probabilities $T$ fixed and learning the emissions associated to its nodes $E$ in the new environment. Further, if we know that the new environment preserves the emission structure, then we can further restrict the learning of $E$, with all the rows of $E$ that correspond to the same observation in the original environment sharing the same parameters. 

The EM algorithm can be used to learn the emission matrix as follows. The E-step recursively computes the forward and backward probabilities and then updates the posterior probabilities. The M-step updates only the emission matrix.

\textbf{E-Step:}
\begin{align}
\tilde{\alpha}(n) &= \left( T(a_{n-1})^{\top}\alpha(n-1) \right) \circ E(x_n) & p_{\alpha}(n) &= \sum_{k=1}^{|Z|} \tilde{\alpha}_k(n) & \alpha(n) =\tilde{\alpha}(n)/p_{\alpha}(n) \\
\tilde{\beta}(n) &= T(a_{n}) \left( \beta(n+1) \circ E(x_{n+1})\right) & p_{\beta}(n) &= \sum_{k=1}^{|Z|} \tilde{\beta}_k(n) & \beta(n) =\tilde{\beta}(n)/p_{\beta}(n) \\
\end{align}

\begin{equation}
\gamma(n) =  \frac{\alpha(n) \circ \beta(n)}{\alpha(n)^\top\beta(n)}
\end{equation}

\textbf{M-Step:}
\begin{equation}
E(j) = \sum_{n=1}^{N} 1_{x_{n} = j} \gamma(n) \oslash \sum_{n=1}^{N} \gamma(n)
\end{equation}
Note that the updates here involve $|Z|\times1$ vectors, where $|Z|$ ($=\sum{N_{cpe}}$) is the total number of hidden states in the model. $T(a_{n})$ corresponds to the transition matrix for the action $a_n$ and is of size $|Z|\times|Z|$, $E(x_n)$ is a column of the emission matrix corresponding to the emission $x_n$, and $1_{x_{n} = j}$ is an indicator function. The forward message is initialized as $\tilde{\alpha}(1) = \pi \circ E(x_1)$; and the backward message is initialized as a vector of all 1s. The emission matrix can be initialized randomly or with equal probabilities for all observations.

When the clone structure is to be preserved, the $\gamma(n)$ term in the E-step is modified as follows. For each observation $j \in \{1, \ldots, N_{\textrm{obs}} \}$, we set the posterior probability for all clones of $j$ to be the same:
\begin{equation}
\bar{\gamma}_k(n) = \sum_{l \in C(j)}\gamma_l(n)\;\; \forall k \in C(j).
\end{equation}

For a given $T$ and $E$, the normalization term $p_{\alpha}(n)$ for the forward messages in the E-step can be used to compute the negative log-likelihood (NLL) of a sequence of observation-action pairs as follows:
\begin{equation}
{\rm NLL} = -\frac{1}{N}\sum_{n=1}^N \log p_{\alpha}(n)
\label{eq:NLL}
\end{equation}

\subsection{Learning the continuous CSCG}
\label{Appendix:continuous_cscg}
When each observation at timestep $n$ is a continuous vector $x_n$ (e.g., an image), we want to maximize the following log-likelihood
\begin{equation}
    \log p(\boldsymbol{x}|\boldsymbol{a};T, \boldsymbol{\mu}) = \log \sum_{z,y} P(z_1)\prod_{n=2}^N P(z_n| z_{n-1}, a_{n-1})  
    \prod_{n=1}^N p(y_n|z_n)\mathcal{N}(x_n |\mu_{y_n}, \sigma^2I)
    \label{eq:logliktolearn}
\end{equation}
w.r.t.~$T$, where $T_{ijk}= P(z_n=k| z_{n-1}=j, a_{n-1}=i)$, and $p(y_n|z_n)$ is 1 for $z_n \in C(y_n)$ and 0 for all others.

This graphical model is tree-structured, so its value can be computed exactly. Parameter optimization can be done via EM, where both the expectation and maximization steps are exact. Rather than learning both $T$ and $\boldsymbol{\mu}$ simultaneously, we find it simpler to proceed in two steps. 

In the first step we fix $P(z_1)=T_{ijk}=1/|Z|$, where $|Z|$ is the number of states in the hidden space, and learn $\boldsymbol{\mu}$ only. The problem thus simplifies to maximizing

\begin{align}
    \nonumber \log p(\boldsymbol{x}|\boldsymbol{a};T=T_\text{uniform},\mu) &= -N\log|Z| + \log \sum_{\boldsymbol{z},\boldsymbol{y}}  
    \prod_{n=1}^N p(y_n|z_n)\mathcal{N}(x_n |\mu_{y_n}, \sigma^2I)\\
    \nonumber 
    &= -N\log|Z| + \sum_{n=1}^N\log \sum_{y_n=1}^K n_C(y_n) \mathcal{N}(x_n |\mu_{y_n}, \sigma^2I)\\
    &= \sum_{n=1}^N\log \sum_{k=1}^K \frac{n_{C}(k)}{|Z|} \mathcal{N}(x_n |\hat\mu_k, \sigma^2I)
    \label{eq:kmeans}
\end{align}
w.r.t. $\boldsymbol{\mu}$. The number of clones of a given centroid $\hat\mu_k$ is denoted by $n_{C}(k)$ (thus, $\sum_{k=1}^K n_{C}(k)=|Z|$). In the last equality we collect all the Gaussians that are known to have the same mean (i.e., all the clones with the same centroid).  The computation is more efficient, since it no longer scales with the total number of hidden states $|Z|$, but only with the number of distinct means $K$.

The astute reader will recognize Eq.~\ref{eq:kmeans} as the log-likelihood of an isotropic Gaussian mixture model, which can be optimized greedily using $K$-means. The centroids $\hat\mu_k$ are the cluster centers and $\frac{n_{C}(k)}{|Z|}$ corresponds to the prior probabilities of each center. In other words, to maximize $\log p(\boldsymbol{x}|\boldsymbol{a};T=T_\text{uniform},\boldsymbol{\mu})$  w.r.t.~$\boldsymbol{\mu}$ we simply need to run $K$-means on the input data $\boldsymbol{x}$ (temporal ordering becomes irrelevant) and then assign a number of clones to each centroid that is proportional to the prior probability of that cluster. When using $K$-means, the value of $\sigma^2$ does not affect the optimization of the centroids. Once the centroids have been chosen, one can set $\sigma^2$ based on the hard assignments from $K$-means, where the maximum likelihood estimate is closed form and simply matches the average distortion. Directly optimizing Eq.~\ref{eq:kmeans} w.r.t. $\boldsymbol{\mu}$ and $\sigma^2$ is also possible using EM (and in principle, more precise), but will be more computationally expensive.

In the second step, we keep $\boldsymbol{\mu}$ fixed as per the previous step, and learn $T$ by maximizing Eq.~\ref{eq:logliktolearn} w.r.t.~$T$ using EM. While it would be possible to learn $\boldsymbol{\mu}$ and $T$ simultaneously in the same EM loop, we find that proceeding in these two steps results in faster convergence without significantly deteriorating the quality of the found local optimum.

Two approaches are possible for the learning of $T$:

\paragraph{Soft-evidence} This would be the vanilla approach of maximizing Eq.~\ref{eq:logliktolearn} w.r.t.~$T$ using EM: in the E-step, we obtain the exact posterior over each $z_n$ by running sum-product inference, and in the M-step we find the maximizing $T$ in closed form for that posterior. This is computationally demanding, since the cost per learning iteration scales with ${\cal O}(N|Z|^2)$.

\paragraph{Hard-evidence} We can apply here the same idea that converts EM for Gaussian mixture modelling into $K$-means (sometimes known as hard-EM). Instead of considering all the centroids as partially \emph{responsible} \cite{verbeek2003efficient} for each sample $x_n$, we can assign all the responsibility to the dominating Gaussian. This is often a very precise approximation, since the dominating Gaussian typically takes most of the responsibility. In turn, this means that at each timestep, only the clones of the centroid that is closest (in the Euclidean sense) can be active. With this change, the cost per learning iteration now scales as ${\cal O}(N n_{C}^2)\ll{\cal O}(N|Z|^2)$, where we have assumed a constant number of clones $n_{C} = n_{C}(k)$ for all centroids for ease of comparison. When doing this, the observations become effectively discrete and learning $T$ reduces to the procedure described in Section \ref{Appendix:learn_em_T}.

In our experiments we use the second approach, which is much more efficient. Putting all together, the final learning procedure can then be summarized as follows:
\begin{enumerate}
  \item Run $K$-means on the training data for a given number of centroids $K$.
  \item Assign to each centroid a number of clones proportional to the prior probablity of that centroid, as obtained by $K$-means.
  \item Vector-quantize the data (train and test) by assigning to each data point the label of the closest centroid.
  \item Learn the action-conditional transition matrix as if this was a discrete CSCG, as explained in \S \ref{Appendix:learn_em_T}.
\end{enumerate}

\subsection{Transfer learning in continuous CSCGs}
\label{Appendix:continuous_cscg_transfer}

In the transfer learning setting we know $T$ for a given environment, and need to learn $\boldsymbol{\mu}$. The principled approach is simple: maximize Eq.~\ref{eq:logliktolearn} w.r.t.~$\boldsymbol{\mu}$ while keeping $T$ fixed. This can be done using EM, where both expectation and maximization will be exact steps. Clone structure can be enforced exactly as well, by tying the parameters of the centroids that correspond to the same clone during the M-step.

The above approach uses soft-evidence and runs into the same computational issues as we discussed in \S \ref{Appendix:continuous_cscg}. If we do not need to keep the same clone structure, we can resort to the same hard-evidence trick to significantly bring down computation time, again from ${\cal O}(N|Z|^2)$ to ${\cal O}(N n_{C}^2)$. The process again reduces to using $K$-means and then doing transfer learning in the discrete space:
\begin{enumerate}
  \item Run $K$-means on data from the new environment. $K$ can be the number of centroids from the environment we are transferring from, but does not need to. The only restriction is that it is not larger than the number of states in $T$.
  \item Vector-quantize the data from the new environment by assigning to each data point the label of the closest centroid.
  \item Learn the emission matrix as if this was a discrete CSCG, as explained in \S \ref{Appendix:learn_em_E} (without preserving the clone structure).
\end{enumerate}

\subsubsection{Schemas for transfer learning}

\textbf{Grounded schema}

A grounded schema is a 3-tuple $S_g\equiv\{T, {\cal C}, E\}$ containing an action-conditional transition tensor $T$, a clone structure ${\cal C}$, and an emission matrix $E$ that respects the clone structure. 

The transition tensor $T$ encodes the transition probabilities, with elements  $$T_{ijk}= P(z_n=k| z_{n-1}=j, a_{n-1}=i)$$
We refer to $T$ as the \textit{schema} portion of the grounded schema.

Latent variables $z$ take values in the range $1,\ldots,|Z|$. This range is partitioned in $K \leq |Z|$ groups of \textit{cloned} states. The group index of a given latent state is given by $1\leq {\cal C}(\cdot) \leq K$. Thus, two latent states $z=j$ and $z=j'$ belong to the same clone group iff ${\cal C}(j) = {\cal C}(j')$.

The emission matrix $E$ encodes the emission probabilities, with elements $$E_{ij} = P(y_n=j| z_n=i)$$ 
An emission matrix is said to \textit{respect the clone structure} if the condition ${\cal C}(j)={\cal C}(j') \implies E_{ij}=E_{ij'}~\forall~i,j,j'$ is satisfied.

\textbf{Ungrounded schema with clone structure}
An ungrounded schema with clone structure is a tuple $S_u\equiv\{T, {\cal C}\}$. It can be derived from a grounded schema $S_g$ by removing the emission matrix $E$. Conversely, an ungrounded schema with clone structure can produce a grounded schema when it is combined with an emission matrix $E$ that respects its clone structure. This process is called \textit{grounding} and corresponds to transfer learning, since the same schema is reused.

\textbf{Ungrounded schema}
An ungrounded schema, or simply a schema, is noted as $T$, and is obtained from a grounded schema in which we have removed both the clone structure ${\cal C}$, and the emission matrix $E$. As before, a schema can be combined with an arbitrary clone structure ${\cal C}$ and an emission matrix $E$ that respects it, to produce a grounded schema. This is an even more flexible form of transfer learning.

\section{Experiment details}
\label{Appendix:experimental_details}

In \cite{George2021-qt}, it was observed that the convergence of EM for learning the parameters of a CSCG can be improved by using a smoothing parameter called the pseudocount. The pseudocount is a small constant that is added to the accumulated counts statistics matrix, which ensures that any transition under any action has a non-zero probability. This ensures that the model does not have zero probability for any sequence of observations at test time. For our experiments, unless otherwise specified, we use a psuedocount of $2 \times 10^{-3}$ for schema learning (learning $T$, with $E$ fixed), and $10^{-7}$ for schema matching (learning $E$, with $T$ fixed).

\subsection{Memory \& Planning Game}
\label{Appendix:MPG}
The game environment is a $4\times 4$ grid of symbols in which the agent can navigate in the four canonical directions by one grid step (up, down, left, right), and collect reward of 1 at a goal location in the grid (Fig.~\ref{fig:mpg_streetlearn}A). Reward is 0 otherwise. Once the agent collects the reward at the current goal, the agent is placed in a new random position and a new goal symbol is sampled to which the agent must navigate to collect the next reward. All grid positions have unique symbols and the symbol-position mapping is randomized at the start of each episode, which lasts for 100 actions. The agent's observation is a tuple of the symbol in its current position and the goal symbol. See \cite{Ritter2020-hu} for more details. We assume knowledge of the collect action function and execute it only when the goal symbol is reached.

We employ 3 different hard coded exploration policies to cover the observations in a new episode: (i) random navigation actions, (ii) random navigation actions but limited to (up, right), and (iii) an optimal set of actions to cover the learned transition graph -- a Hamiltonian cycle that repeats the sequence \{up, up, up, right\} four times. For a new episode, we take actions based on one of these policies, learn the emission matrix $E$ at every step, and plan to reach the current goal using the current estimate of $E$. Planning a path from the current position to a goal position is achieved via Viterbi decoding in the CSCG. If the probability of reaching the goal is above a certain threshold, we find the series of actions that lead to the goal. 

In the beginning, the agent is uncertain about both its own position and the goal position, so it executes actions based on one of the hard-coded policies until we are confident of reaching the goal state from the current state above a probability threshold. If the current goal symbol is not yet observed in this episode, we execute actions based on the policy, otherwise we compute the probability of our planned set of actions reaching the goal symbol. As we navigate, the estimate of $E$ becomes better and the plans are more likely to succeed. Our planning algorithm takes both these uncertainties into account. We evaluated for a total of 100 episodes ($10^4$ steps) and the average reward stays the same after the 9th  episode once we learn the schema. Average reward per episode are reported for 100 episodes from 10th episode onwards i.e., after learning.

\subsection{One-Shot StreetLearn}
\label{Appendix:streetlearn}
In each episode of this task, the agent is placed in a new neighborhood of a city and needs to move to a goal location and direct itself to the target, specified by the goal street view image, to collect the reward. The agent can take one of four actions: move forward, turn left, turn right, collect reward. After every reward collection, the agent is placed in a new location with a new heading direction and provided a new goal street view image. Each episode lasts 200 steps and a new neighborhood is sampled for each new episode. The agent needs to collect as many rewards as possible during each episode. The StreetLearn environment represents the agent's perception, a street view image, as a unique compressed string that we use as our agent's observation. For the CSCG agent, we follow an explore and exploit strategy. The agent first navigates every action from every observation it encounters while collecting the rewards as it encounters the goals. This is guided exploration to cover every possible edge in the transition graph. A CSCG model is then learned from the sequence of observations and actions experienced in this episode. For any subsequent step, we find the closest path from the agent's current observation to the goal observation and execute the actions, re-planning at every step to account for any mismatch between the model expectation and reality.

\subsection{Schema matching}
\label{Appendix:room_schema_recognition}

\subsubsection{Schema matching in simulated 3D environments}
\label{Appendix:agix}

In our simulated 3D environments, the agent receives egocentric observations that are $64 \times64$ RGB images. The agent's action space is discretized to 3 egocentric actions: go forward by $1$m, rotate $90^\circ$ left, and rotate $90^\circ$ right. We convert the input RGB images into categorical observations using a $K$-means vector quantizer.

\begin{figure}[ht!]
\begin{center}
\includegraphics[width=\linewidth]{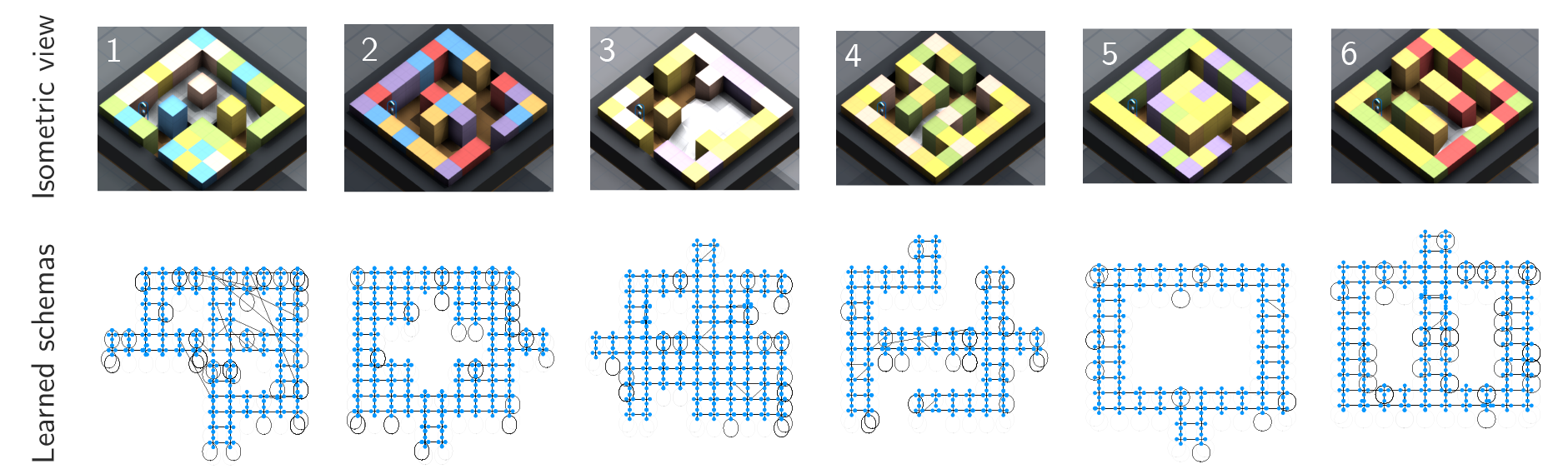}
\end{center}
\caption{ The top row shows the isometric views of example 3D simulated environments used in our schema matching experiments. The bottom row shows the transition graphs of CSCGs trained on the respective 3D environments. Note that each location in the room is represented by four nodes in the transition graph corresponding to the four possible heading directions of the agent.
}
\label{fig:agix_cscg_models}
\end{figure}

\begin{figure}[ht!]
\begin{center}
\includegraphics[width=\linewidth]{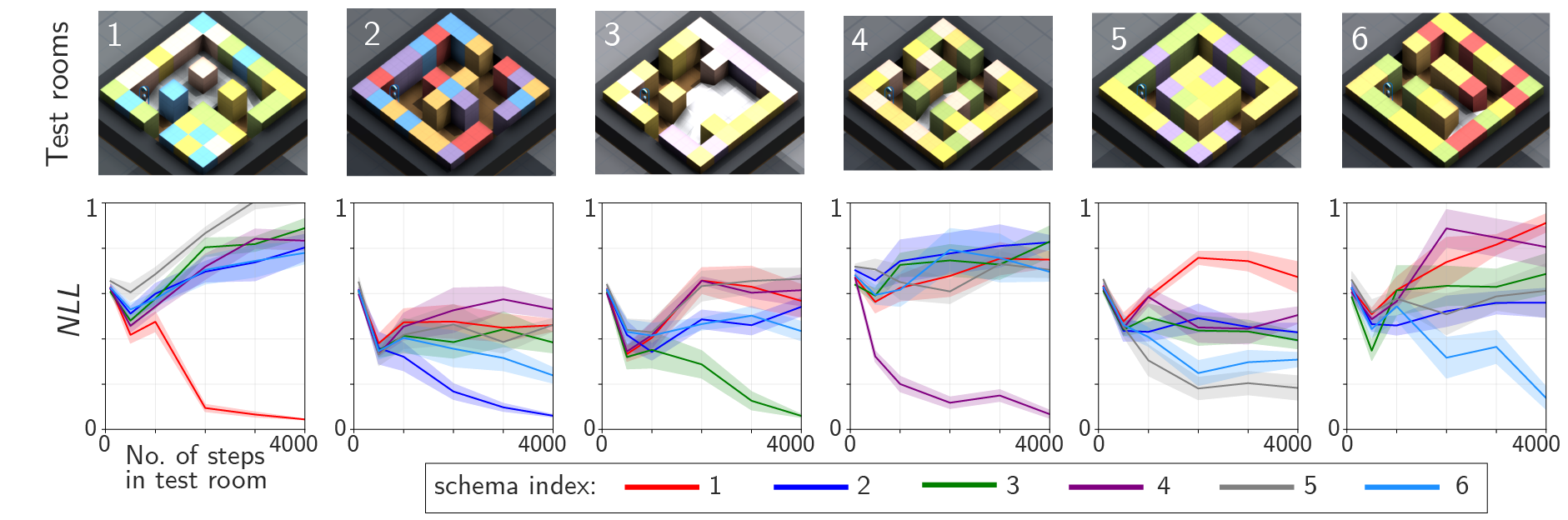}
\end{center}
\caption{Each column shows an example test room (top) and the negative log-likelihood (NLL) of observations from that room under different schemas (bottom). Error bars are SEM. Note that the correct schema was identified (lowest NLL) in all 6 test rooms.}
\label{fig:agix_schema_recognition}
\end{figure}

For the schema matching experiments with 3D environments, we used the six rooms shown in Fig. \ref{fig:agix_cscg_models}. These were constructed with a variety of features, including ramps and platforms that introduce one-way transitions (e.g. dropping off of a ledge). For each of these environments, we trained a CSCG using the following steps. First, we collected action-observation pairs from a random walk of length $100,000$ steps. Then, we clustered the egocentric observations from the walk using a $K$-means quantizer, with $K=50$. We used EM to train a CSCG on these discretized observations and actions as described in \S\ref{Appendix:learn_em_T}. The learned transition graphs for the six example rooms are also shown in Fig. \ref{fig:agix_cscg_models}. Note that learning is not always perfect. Some of the transition graphs have spurious edges. These imperfections, however, do not affect our schema matching results.

Fig. \ref{fig:agix_schema_recognition} shows the schema recognition results in these 3D rooms. Given observations from a test room and a hypothesized schema (ungrounded and without clone structure), we learn the emission bindings as described in \S\ref{Appendix:continuous_cscg_transfer}. We compute the negative log likelihood of the observations according to Eq.~\ref{eq:NLL} in \S\ref{Appendix:learn_em_E}. We observed that it takes typically between $500$ to $1000$ steps to correctly identify the matching schema in each room, although this is higher in a couple cases. In particular, disambiguation of schemas 5 and 6 required closer to 2000 steps.

\begin{figure}[ht!]
\begin{center}
\includegraphics[width=0.95\linewidth]{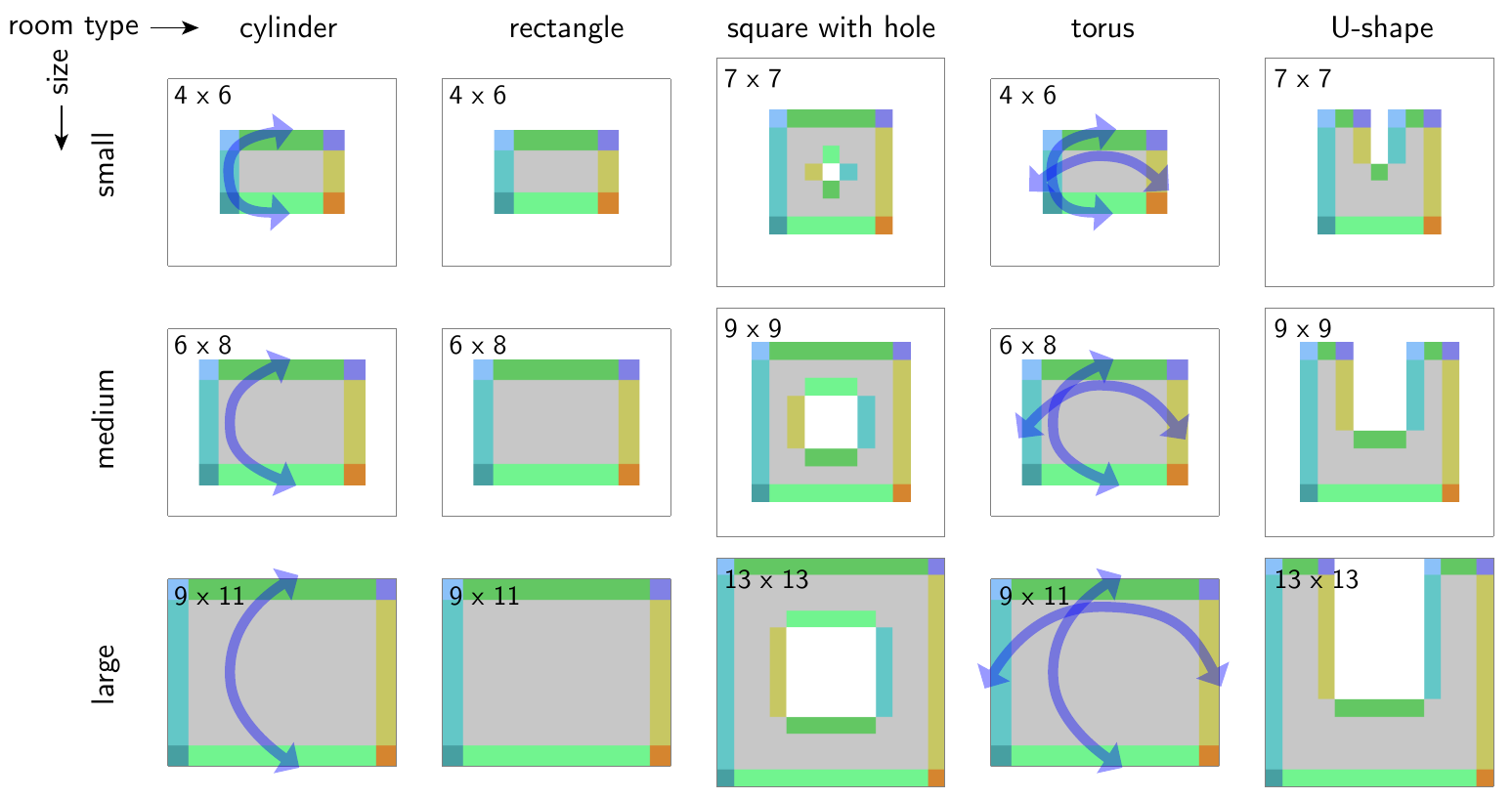}
\end{center}
\caption{Rooms of different types and sizes used in the schema matching experiments with size variations.}
\label{fig:all_shape_rooms}
\end{figure}

\subsubsection{Schema matching in 2D rooms with size variation}
\label{Appendix:schema_matching_2D_rooms}
We generated rooms of five different types, and three size variations per type, as shown in Fig.~\ref{fig:all_shape_rooms}. We selected the room and barrier dimensions such that the rooms have similar number of accessible states. For the torus and cylinder rooms, we use the same observation map as in the rectangular room case. However, actions wrap around the top-bottom edges for the cylinder, and the top-bottom and left-right pair of edges for the torus, as indicated by the blue arrows in Fig.~\ref{fig:all_shape_rooms}. We learned schemas on the medium size variation of these room types. For each training room, we used action-observation sequences from a random walk of length $50,000$ steps to train the respective CSCG.

For evaluating schema matching, we used test rooms of all three size variations. Importantly, we generated the test rooms by permuting the observations from the original room. As a consequence, at test time, our model would have to relearn the observation mapping. 

\begin{figure}[ht!]
\begin{center}
\includegraphics[width=\linewidth]{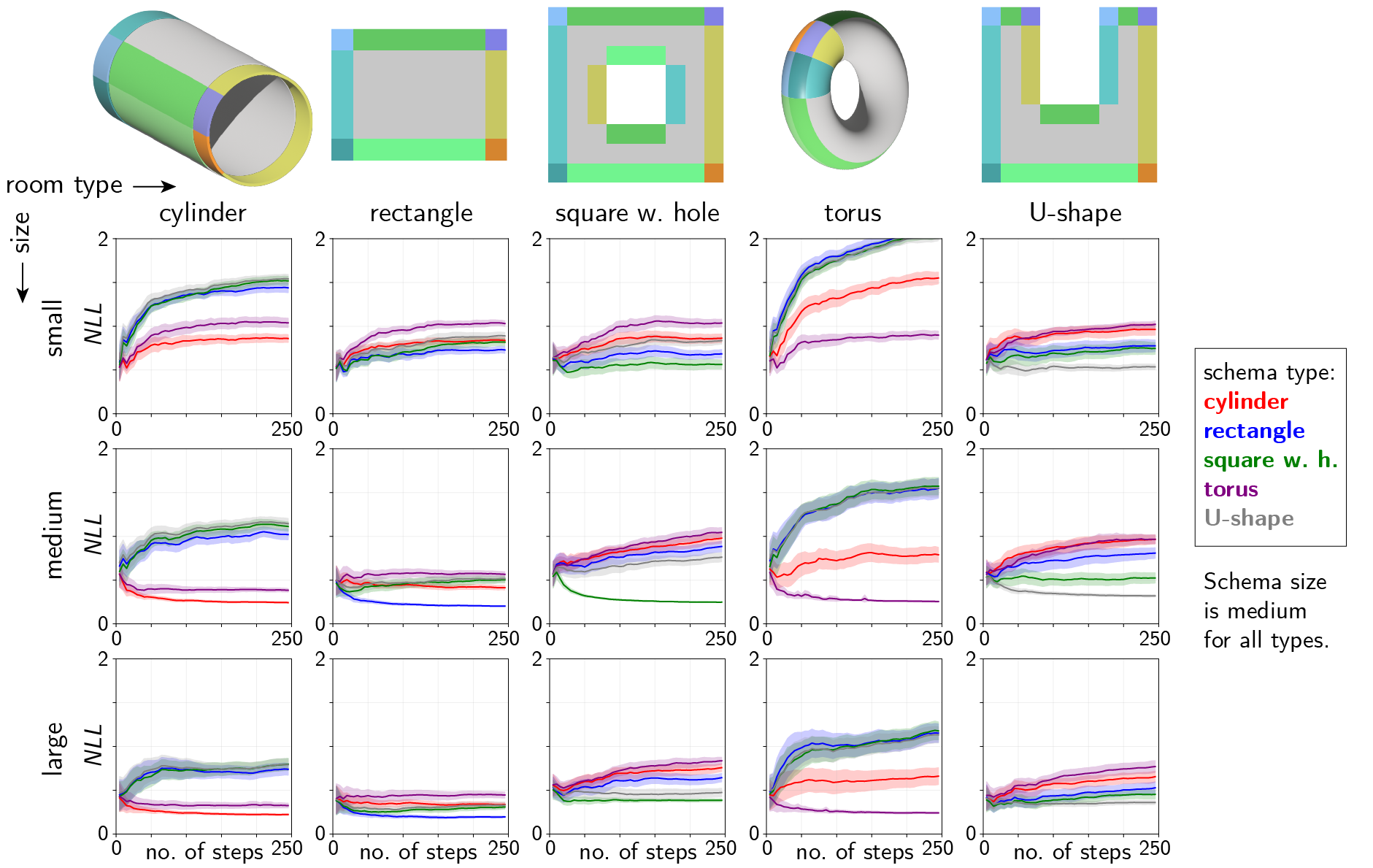}
\end{center}
\caption{Schema matching in novel environments with size and observation variations. We measure the negative log likelihood (NLL) of observation sequences in a given room under different schemas as the agent does random walks. The schema with the least negative log likelihood (conversely the highest likelihood) is considered the best match. Each panel corresponds to a room of type and size indexed by the column and row headers, respectively. Schemas are learned in medium sized rooms and tested on two other size variations. Error bars are 95\% CI of the SEM.
}
\label{fig:schema_matching_no_clone_structure}
\end{figure}

In Fig. \ref{fig:schema_matching_no_clone_structure}, we show the schema matching performance with the use of clone-structure. We observe that it takes at most $95$ steps to identify the correct schema for all test rooms when using clone structure. We obtained these results using the paired T-test over 25 random walks, per test room, and a p-value threshold of $0.05$. We also evaluated the schema matching performance without clone structure and found that it takes at most $255$ steps to find the best matching schema when the clone structure is ignored.

\begin{figure}[ht!]
    \begin{center}
    \includegraphics[width=0.9\linewidth]{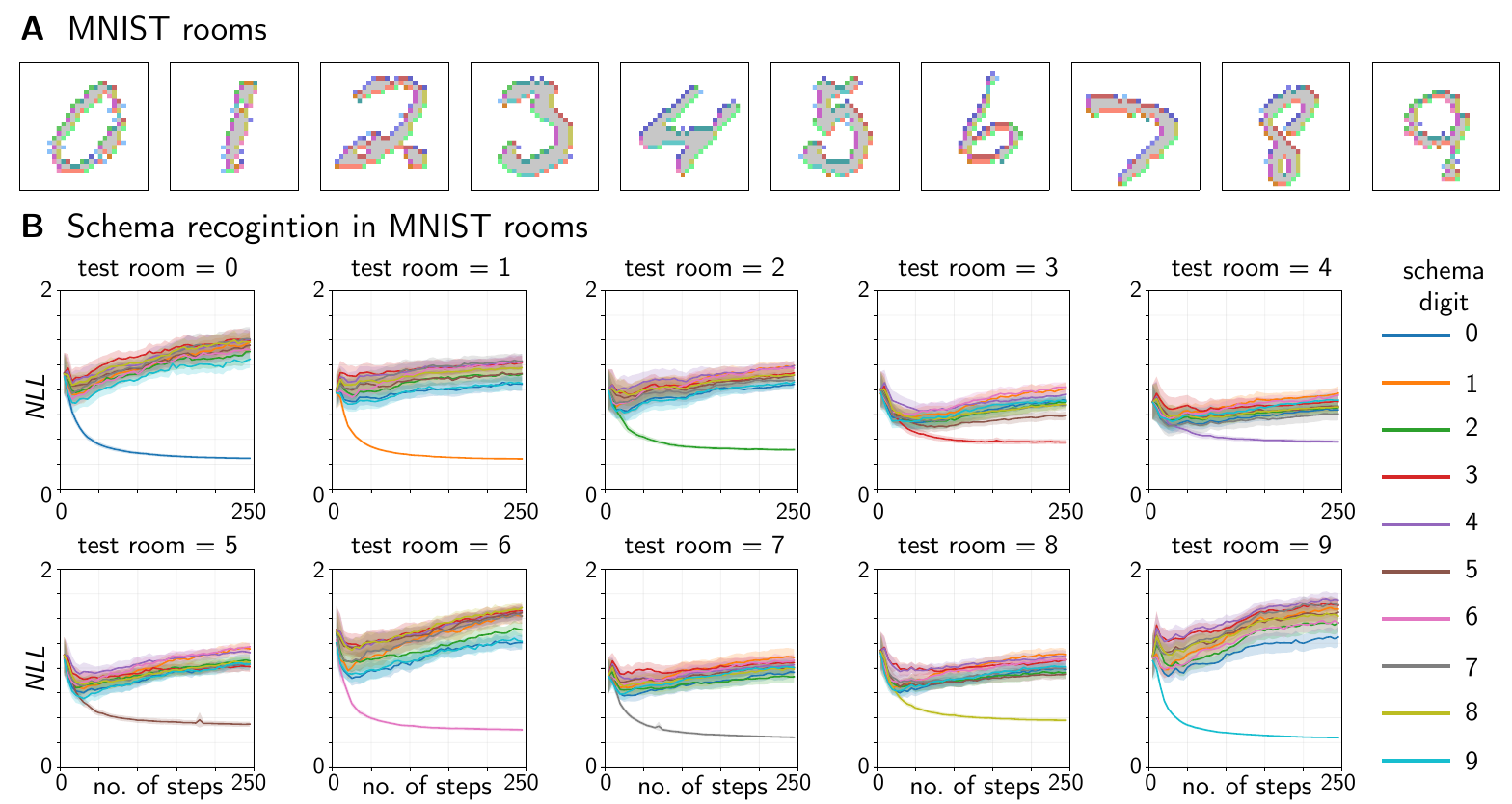}
    \end{center}
    \caption{{\bf A}. MNIST rooms constructed using the binarized-MNIST dataset, with the observations mapped as colors.
    {\bf B}. Negative log likelihood (NLL) of observation sequences from a given MNIST room under different schemas.
    Error bars are 95\% CI of the SEM.}
    \label{fig:appendix-mnist_schema_recognition}
\end{figure}

\subsubsection{Schema matching in MNIST digit rooms}
\label{Appendix:mnist_experiments}
In this experiment, we used a set of ten digits from the binarized MNIST dataset as our room layouts as they provide an interesting variety of shapes and topologies. All the foreground pixels are treated as accessible locations; background pixels are treated as obstacles. Observations at each pixel are sampled in a similar way to the 2D rooms with aliased interiors (Fig.~\ref{fig:appendix-mnist_schema_recognition}A). We learn schemas for 10 selected  MNIST-digits using action-observation pairs from random walks of length $100,000$ steps in each room. 
We repeated our schema matching procedure as before in this new set of rooms by keeping the room structures constant but with new observations. Fig.~\ref{fig:appendix-mnist_schema_recognition} shows the negative log likelihoods of observation sequences in test rooms under different schemas. The correct schema was identified in at most $50$ steps in all test rooms.

\subsection{Schema matching in composed environments}
\begin{figure}[ht!]
\begin{center}
\includegraphics[width=1\linewidth]{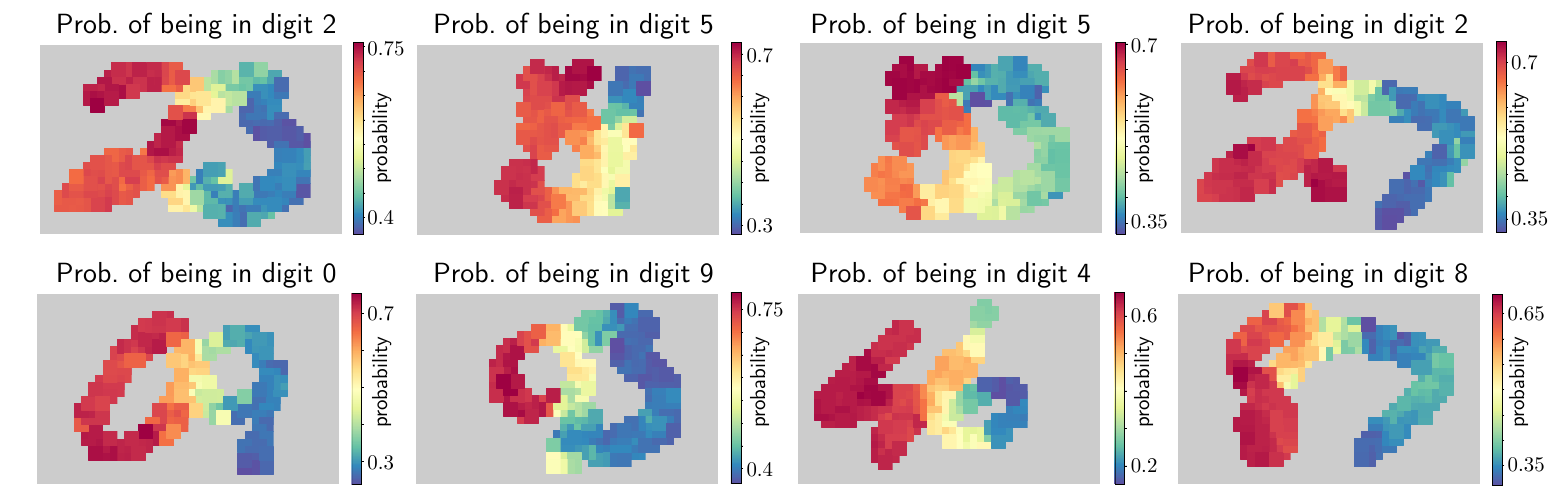}
\end{center}
\caption{Each heatmap represents the probability of an agent being in one of two schemas at each location in an MNIST composed room, averaged over all instances of the agent being at that location during a random walk of length $20,000$ steps.
}
\label{fig:appendix-mnist_joint_recognition}
\end{figure}

To demonstrate how schemas could be used in more complex environments composed of multiple schemas, we constructed test rooms composed of pairs of overlapping MNIST digits. As an agent randomly explores a composed room, we compute the likelihood of a sliding window of action-observation pairs under each schema. Note that we use a sliding window here, instead of the entire history of observations, because we are interested in identifying schema the agent is currently in. The probability of being in a particular schema can then be computed using the softmax of the log-likelihoods. The details of schema matching in composed MNIST rooms is presented in Algorithm \ref{alg:composed_schema_selection}. 
Heatmaps in Fig.~\ref{fig:appendix-mnist_joint_recognition} show the estimated probability of being in one of the two schemas at every location in a composed room, as the agent performs a random walk of length $20,000$ steps and sliding window of length $200$ steps. We also compute the accuracy of identifying the correct schema at all locations by thresholding these probabilities. The average accuracy of identifying the correct schema at all locations in all composed test rooms is $88.4 \pm 0.9\%$. See Table \ref{table:mnist_composition} for results on individual composed rooms.

\begin{algorithm}[ht!]
\caption{Schema selection for composed rooms}
\label{alg:composed_schema_selection}
\textbf{Input} List of schemas $S_1, \ldots S_H$, observations $\boldsymbol{x}$ and actions $\boldsymbol{a}$ of length $N$ in the test room, window length $w$.
\begin{algorithmic}[1]
\State $n \gets w$
\While{$n <= N$}
    \State Estimate $E_j$ using $x_{n-w},\ldots,x_n$, $a_{n-w},\ldots,a_n$ for each schema, $j \in \{1,\ldots, H\}$
    \State Compute $L_{n,j}$: log likelihood of $x_{n-w},\ldots,x_n$ under schema $j$
    \State Compute $p_{n,j} = \mathrm{softmax}(L_{n})_j$ : the probability of being in schema $j$ at time step $n$. 
    \State Select schema with highest $p_{n,j}$.  
\EndWhile
\end{algorithmic}
\end{algorithm}

\begin{table}[ht!]
\centering
\begin{tabular}{ |c||c|c|c|c|c|c|c|c| } 
 \hline
 Digits & 2, 3 & 5, 1 & 5, 3 & 2, 7 & 0, 9 & 9, 3 & 4, 6 & 8, 7 \\ 
 \hline \hline
 Acc. & ${87.6}\pm{4}$ & ${84.3}\pm{3}$ & ${91.1}\pm{3}$ & ${92.5}\pm{2}$ &  ${87.9}\pm{3}$ & ${85.2}\pm{4}$ & ${85.9}\pm{3}$ & ${90.7}\pm{4}$ \\ 
 \hline
\end{tabular}
\caption{Accuracy (Mean $\pm$ Standard deviation, in \%) of the schema predictions as an agent moves across various locations in composed MNIST rooms. We use the ground truth locations to compute this accuracy, which is averaged over five trials of random walks in each room. Note that for the composed room experiments, the agent gets two schemas as input.}
 \label{table:mnist_composition}
\end{table}

\subsection{Learning using schema compositionality}
\label{Appendix:schema_stitching}
We generated novel environments by composing rooms with known schemas. For each individual schema, we specify its frontiers as potential exits and entries: (i) list of (state, action) tuples that specify how the agent can exit the schema, and (ii) list of entry states from which an agent can enter the schema. We pre-specify schemas and frontiers for our experiments, but schemas and their frontiers can potentially be learned from training data (e.g. as repeating motifs on a set of training environments using graph clustering methods and metrics such as betweenness centrality). Once the schemas and frontiers are specified, a novel composed environment is thus a combination of different regions corresponding to known schemas that are connected by their frontiers.

\begin{algorithm}
\caption{Learning a new CSCG model with schema compositionality}\label{alg:schema_stitching}
\textbf{Input} List of schemas $S_1, \ldots, S_H$, frontiers $FSA_1, \ldots, FSA_H$, observations $\boldsymbol{x}$, actions $\boldsymbol{a}$, and pseudocount $\alpha$\\
\textbf{Output} CSCG model of the room as $T$ and $E$.
\begin{algorithmic}[1]
\State{$T\gets {\rm block\ diagonal\ matrix}(S_1, \ldots S_H)$}
\Comment{Initialize $T$}
\For{$i$ in $1, \ldots, H$}
\Comment{Add frontier connections}
    \For{exit frontier tuple $(s, a)$ in $FSA_i$}
    
        $T[a, s, s_k] \gets \frac{1}{|\lbrace s_k \in FSA \setminus FSA_i\rbrace|}\,\, \forall{s_k \in \lbrace FSA \setminus FSA_i\rbrace}$
    \EndFor
\EndFor
\State{$E \gets {\rm Uniform\ distribution}$}
\Comment{Initialize $E$}
\State{Use EM with fixed $T$ to learn $E$ from $(\boldsymbol{x}, \boldsymbol{a})$}
\Comment{Learn $E$ with initial $T$ fixed}
\State{Use EM with $E$ fixed to learn $T$ with pseudocount $\alpha$}
\Comment{Learn $T$ with learned $E$ fixed}
\State{Use Viterbi learning with current $E$ fixed to refine $T$}
\Comment{Refine $T$}
\end{algorithmic}
\end{algorithm}

To learn a model of the composed environments, we use the observations and actions collected during random walks. We first create a new CSCG model with a joint transition tensor composed of known schemas as a block-diagonal tensor, in which each block corresponds to a copy of one schema. We then connect exit frontier states of a given schema copy to every other entry frontier (state) in every other schema with a uniform probability of such transitions conditioned by the corresponding frontier actions for those exit states. See Fig. \ref{fig:schema_stitching} (main text) for examples of graphs representing such joint transition tensors. The procedure for learning the emission matrix and the joint transition tensor using action-observation pairs in a composed environment is described in Algorithm~\ref{alg:schema_stitching}.

For the schema composition experiment, we used schemas for the six environments that were used in the 3D schema matching experiment (\$\ref{Appendix:agix}, and tested learning in a larger environment that was a composition of four of these rooms. Because this complex environment, with one-way transitions in some places, was difficult to cover effectively with a random walk, we opted to use a more efficient walk generation procedure to obtain data for learning. This procedure utilized ground-truth position knowledge to navigate, on every step, towards the nearest unexplored (or least explored) transition. Using this procedure, we were able to cover the entire composed environment in  about $10,000$ steps and generate data that was relatively balanced in its coverage of the different parts of the environment for a wide range of walk lengths.

We first collected efficient walk data of $50,000$ steps for the multi-room environment, and clustered image observations using $K$-means with $K=80$. We constructed exact schemas for each of the six smaller rooms, and used these to construct the initial joint transition tensor with six blocks, connecting schema exit states to entrance states of every other schema (always for the forward action only). The joint transition tensor was then used as the starting point for the learning algorithm (Algorithm \ref{alg:schema_stitching}), which involved learning both $E$ and $T$ for the multi-room environment. 

We experimented with learning over different random walk lengths, from $1,000$ up to $30,000$. These walks were subsets of the full efficient walk data that was collected. We computed mean NLL across 10 test random walks, each of length $10,000$ steps, extracted from a separate random walk dataset that was not used for training. As seen in Fig. \ref{fig:schema_stitching}B, with $5,000$ steps, we observed several spurious edges in the learned composite transition graph. The NLL of test data using this model was 0.73. With $10,000$ steps, we were able to perfectly recover the joint transition graph. The NLL of test data using this learned model was 0.17. In contrast, learning the CSCG for the multi-room environment without schemas, with only $10,000$ steps (test NLL = 4.27) or even $30,000$ steps (test NLL = 1.21) was not feasible. A comparison of the NLL of test data using CSCGs trained with and without schemas as a function of training data size is shown in Fig. \ref{fig:schema_stitching}B.

\begin{figure}[ht!]
\begin{center}
\includegraphics[width=\linewidth]{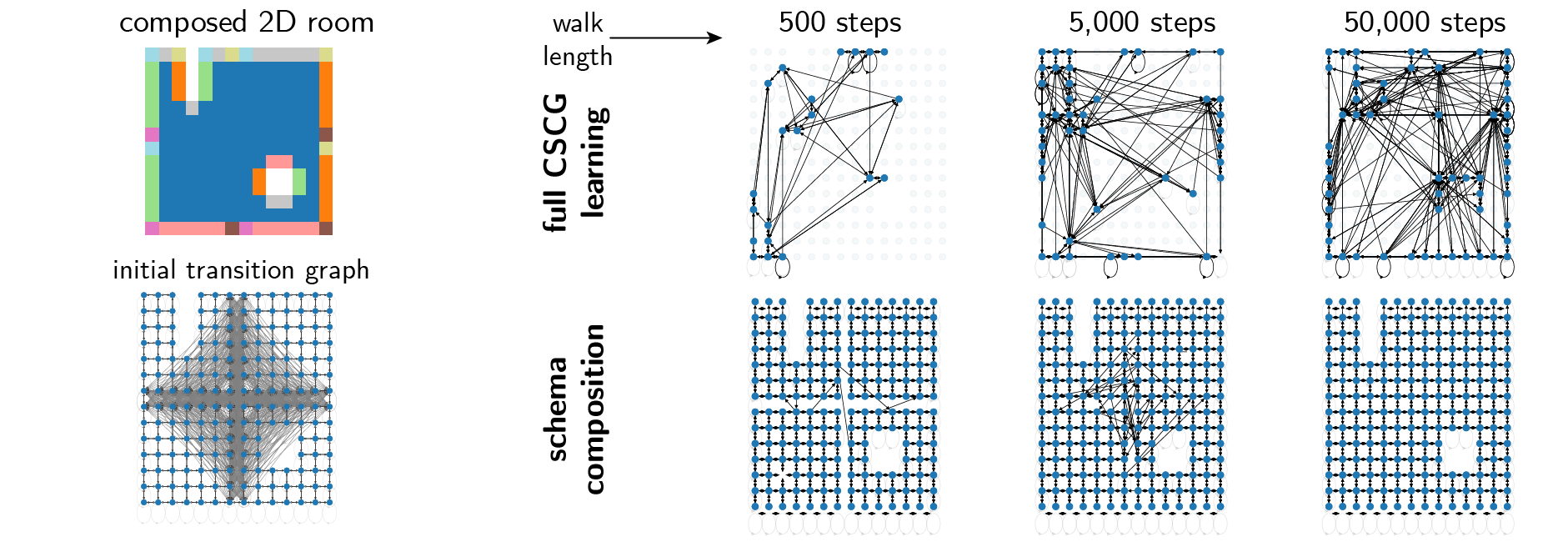}
\end{center}
\caption{Visual comparison of CSCG training with (schema composition) and without (full CSCG learning) schemas, as a function of the length of the training sequence, in an example composed 2D room.}
\label{fig:appendix-2D-composition}
\end{figure}

We also performed the schema composition experiment using 2D rooms from \S\ref{Appendix:schema_matching_2D_rooms}. In Fig. \ref{fig:appendix-2D-composition}, we visualize the models learnt with and without schemas for different walk lengths.

\begin{figure}[ht!]
\begin{center}
\includegraphics[width=\linewidth]{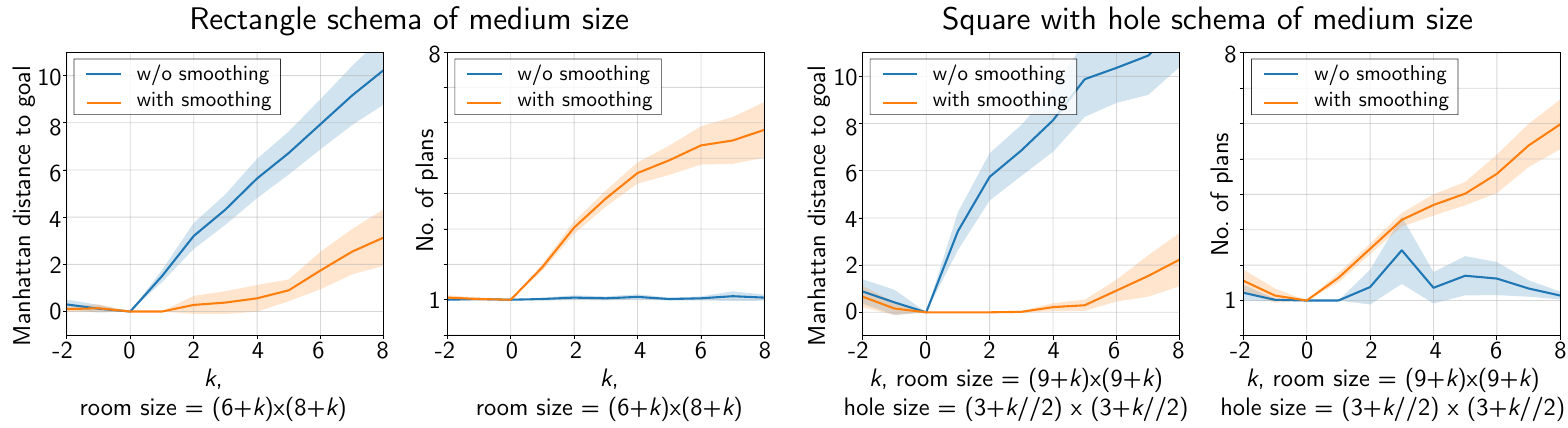}
\end{center}
\caption{
Shortcut finding performance in rectangle and square with a hole room layouts as measured by the Manhattan distance from the goal and the number of plans/re-plans to reach the goal, as a function of room size. Error bars are 95\% CI of the SEM.}
\label{fig:appendix-shortcut_planning}
\end{figure}

\subsection{Rapid inference and planning using schemas}
\label{Appendix:shortcut_finding}

Rapid schema matching and binding enables planning in novel environments with limited experience. We evaluate planning to reach goals using schemas in test rooms with different size and shape variations. We considered two test room types -- rectangle and square with a hole. The schemas correspond to the medium size version of these rooms (same as in \S\ref{Appendix:schema_matching_2D_rooms}). We add an additional smoothing term $\lambda \max(T)$ to the diagonal elements of the transition tensors of each schema. This diagonal smoothing term controls the self-transition probability at each node for each action. This is especially important for generalization to size and shape variations as the previously learned schemas can be rigid. For these experiments, we used $\lambda=0.2$.

For each room size and shape, we generate a sequence of random actions and repeat every action 3 times. We select a starting location corresponding to a unique observation in the room and execute these actions up to 200 steps and collect the observations. Same action repetition creates trajectories where the agent walks farther from the start position with less opportunity to fully explore the surrounding area. The agent is then tasked with going back to the start location of the walk by planning a shortest path back to the start location from the current location. Based on the matched schema, we first learn the emission bindings from the observation-action pairs. We then use the grounded schema (i.e., schema + learned emissions) to Viterbi decode the the current state and the start state, and plan the path back to start state in the model using max-product message-passing. After executing every action of this plan, we collect the observations and re-plan when the current plan fails to take us the goal state. We repeat this for $50$ such random walks in each test room. Fig. \ref{fig:appendix-shortcut_planning} shows the shortcut planning performance for the two example room types.

\end{document}